\documentclass{article}
\usepackage[main, final]{neurips_2026}
\usepackage{amsmath}
\usepackage{amssymb}
\usepackage{algorithm}
\usepackage{algpseudocode}
\usepackage{graphicx}
\usepackage{placeins} 
\usepackage{xcolor}
\usepackage{xurl}
\usepackage{booktabs}
\usepackage{multirow}
\usepackage{enumitem}
\usepackage{hyperref}       

\title{Empirical Bayes Flow Matching for\\Continuous Cryo-EM Heterogeneity}

\author{%
  Daniel Aibinder \\
  University of Haifa \\
  \texttt{daibinde@campus.haifa.ac.il} \\
  \And
  Azmi Haider \\
  University of Haifa \\
  \texttt{ahaide03@campus.haifa.ac.il} \\
  \And
  Dan Rosenbaum \\
  University of Haifa \\
  \texttt{danro@cs.haifa.ac.il} \\
}

\begin{document}

\maketitle
\begin{abstract}
We address the problem of learning generative priors over high-dimensional latent variables from indirect, noisy observations, with a focus on continuous heterogeneity in cryo-electron microscopy (cryo-EM). In this setting, 3D atomic structures $x \in \mathbb{R}^{3N_a}$ are never directly observed and must be inferred from 2D projection images $y$ generated by a complex, non-linear forward operator. We propose Empirical Bayes Flow Matching (EB-FM), a method that learns a flow matching prior $p_\theta(x)$ directly from observations using a Monte Carlo stochastic approximation Expectation-Maximization  framework. EB-FM alternates between an E-step that performs approximate posterior sampling from $p_\theta(x \mid y)$ via guided flow-based generation while maintaining persistent latent estimates through stochastic approximation, and an M-step that updates the flow model using these stabilized latent variables as training targets. This approach eliminates the need for pretraining on clean data and naturally accommodates non-linear forward models. We validate our method on controlled inverse problems using two-moons and MNIST data, and demonstrate its effectiveness on simulated cryo-EM data with known poses, showing that EB-FM can recover continuous distributions of protein conformations directly in atomic-coordinate space.
\end{abstract}

\begin{figure}
\centering
\includegraphics[width=1\linewidth]{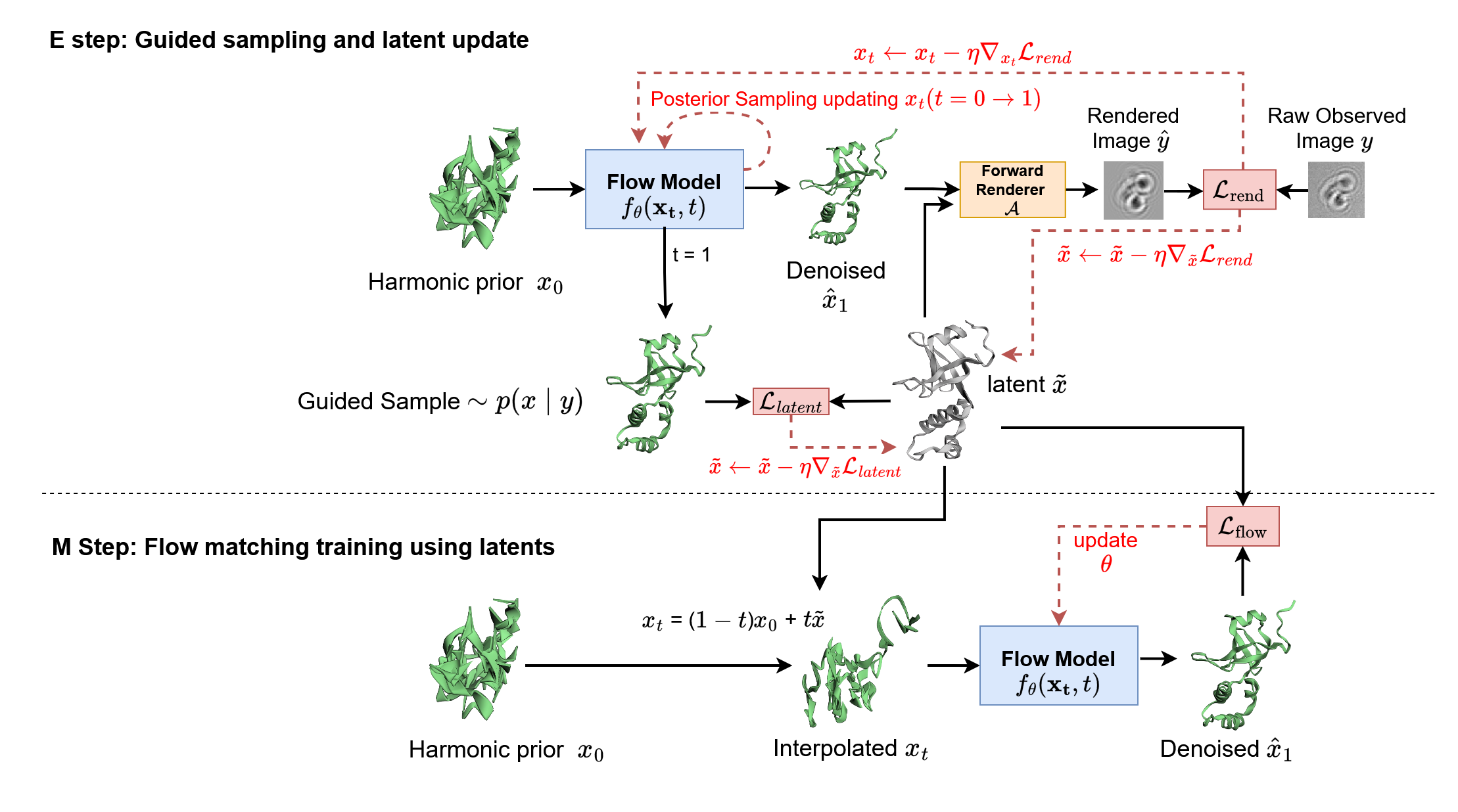}
\caption{Overview of EB-FM, illustrated for the cryo-EM setting. Each observation $y_i$ is associated with a persistent latent $\tilde{x}_i$ that tracks the posterior mean. In the E-step, the current flow matching prior $p_\theta(x)$ generates a posterior sample $\hat{x}_i$ via gradient-based guidance through the differentiable forward operator $\mathcal{A}$, and $\tilde{x}_i$ is updated by EMA. In the M-step, the prior is trained by conditional flow matching on the persistent latents $\{\tilde{x}_i\}$ in place of clean data.}
\label{fig:overview}
\end{figure}

\section{Introduction}

In the natural sciences, quantities of interest are typically not directly observed but inferred from indirect measurements. This inverse-modelling task is central to experimental science, and its success depends on how faithfully we can represent the distribution of the underlying quantities of interest.

Inferring protein structure using cryogenic electron microscopy (cryo-EM) is a prime example of this task \cite{frank2002single, egelman2016current, nogales2015cryo}. Proteins carry out their biological functions as dynamic molecules, adopting multiple conformational states that mediate essential cellular processes such as transport and signalling~\cite{frauenfelder1991energy, motlagh2014ensemble}. Cryo-EM is uniquely positioned to observe this structural heterogeneity: a biological sample is flash-frozen, preserving the ensemble of conformations present in solution, and an electron beam then produces thousands of noisy 2D projection images of the 3D structures, each in a random, unknown orientation~\cite{frank2002single, egelman2016current, nogales2015cryo, vulovic2013image}. In contrast to models like AlphaFold \cite{jumper2021highly,abramson2024accurate}, which primarily target high confidence static structures, cryo-EM data directly reflects the full distribution of conformations that a protein occupies, and recovering this  distribution is essential for understanding function \cite{GIRI2023102536}. The task is therefore not to reconstruct a single 3D structure, but to infer a distribution $p(x)$ over 3D atomic configurations from indirect observations. This can be framed as an Empirical Bayes problem~\cite{robbins1964empirical,casellaIntroductionEmpiricalBayes1985,efronTwoModelingStrategies2014}.

Classical cryo-EM reconstruction pipelines \cite{scheres2012relion, punjani2017cryosparc} target single consensus structures and handle heterogeneity only through discrete clustering. Recent deep-learning methods (cryoDRGN~ \cite{zhong2021cryodrgn}, 3DFlex~\cite{punjani20233dflex}, cryoFIRE~\cite{levy2022cryofire}, cryoSPHERE~\cite{ducrocq2025cryospheresingleparticleheterogeneousreconstruction}, mixture-of-neural-fields~\cite{levy2024mixtureneuralfieldsheterogeneous})  learn continuous latent representations of conformational variability, but all share a structural limitation: they operate in a two-stage regime in which the 2D projections are first used to reconstruct 3D density / voxel grids, and atomic models are fit to those densities only as a separate downstream step. 
This decoupling of the generative model from the underlying atomic physics prevents the use of strong
structural priors: the density stage has no access to atomic-coordinate priors, and the atom-fitting stage no longer has access to the raw observations. This matters most when the observations are under-determined and have low signal-to-noise ratios, precisely the regime where a strong atomic prior should help the most. The exception in the literature is \cite{rosenbaum2021inferringcontinuousdistributionatom}, which operates directly in atomic-coordinate space and couples an atomic-coordinate prior to the raw image observations using a variational autoencoder. 
We follow this direction, incorporating modern generative-modelling approaches for the prior.

Diffusion \cite{sohl2015deep, ho2020denoisingdiffusionprobabilisticmodels, song2021scorebased} and flow matching \cite{liuFlow2023,lipman2023flowmatchinggenerativemodeling, albergo2025stochastic} models have become the dominant framework for learning complex high-dimensional distributions, with striking successes in molecular and structural biology \cite{jing2023eigenfold, ingraham2023chroma, watson2023rfdiffusion,jing2024alphafoldmeetsflowmatching,abramson2024accurate}. 
Recent work has demonstrated that pre-trained diffusion priors can be successfully leveraged to solve inverse problems directly in protein space. Notably, Levy et al.~\cite{levy2024solving} and CryoBoltz~\cite{raghu2025cryoboltz} successfully guide pre-trained diffusion models to fit heterogeneous cryo-EM density maps at inference time. However, utilizing a pre-trained prior is not feasible for \textit{de novo} structural discovery,  where no clean atomic structures of the target distribution are available and the training data consists only of indirect, noisy projections $\{y_i\}$. The central question is therefore how to train a modern generative prior when the data is itself the output of the inverse problem we are trying to solve.

We propose Empirical Bayes Flow Matching (EB-FM), a method that learns a flow matching prior over atomic structures from indirect observations, by iteratively inverting the forward model during training. Like \cite{rosenbaum2021inferringcontinuousdistributionatom}, we operate in atomic-coordinate space, but we depart from its VAE framing in two important ways. First, we replace the variational-autoencoder prior with a flow matching model, which can capture complex multi-modal structural distributions. Second, instead of an amortised encoder, we maintain persistent per-observation latents $\{\tilde x_i\}$ that are inferred via optimization during the model training. This is formulated as a Monte Carlo stochastic approximation Expectation-Maximization algorithm~\cite{delyon1999convergence}. We optimize the marginal likelihood of the observations by alternating between an E-step, where each latent $\tilde x_i$ is refined via deterministic approximate posterior sampling using  the current prior and the observation $y_i$, and an M-step, in which the flow matching prior is retrained on the stabilized latent ensemble $\{\tilde x_i\}$. 

Several lines of work train generative priors directly from corrupted observations rather than clean data, including AmbientDiffusion~\cite{darasAmbient2023} for structured measurement corruption. Most relevant to our setting,  DiEM~\cite{rozet2024learning} proposes a similar empirical Bayes scheme with a diffusion prior, but under two restrictions that prevent its application in our setting. First, DiEM is formulated exclusively for linear Gaussian forward models $p(y \mid x) = \mathcal{N}(y \mid Ax, \Sigma_y)$, for which the intermediate-state marginal likelihood admits a closed form via moment matching. Our cryo-EM forward model is structurally non-linear as the 3D atomic coordinates enter the projected image through a sum of radial-basis exponentials. Second, DiEM's E-step relies on solving a large linear system via a Conjugate-Gradient method at every discretization step. This requires evaluating Vector-Jacobian Products through the generative network, which becomes computationally prohibitive in the case of high dimensional observations. EB-FM bypasses these limitations, targeting non-linear differentiable forward models with a tractable E-step. 

We first validate EB-FM on controlled benchmarks using 2D two-moons data and MNIST, where we can ablate design choices. We then apply it to the cryo-EM task of inferring heterogeneous protein structure, with data generated by molecular-dynamics and electron-microscopy simulators~\cite{vulovic2013image, rullgaard2011simulation}. We adopt known poses as a simplifying assumption that isolates the EB-FM contribution from the joint pose-estimation problem. While unknown poses and mismatch between simulated and real data remain substantial open challenges, EB-FM provides the methodological foundation on which those extensions can be built. Our contributions can be summarized as follows: 
\begin{enumerate}[nosep, leftmargin=20pt]
\item We develop empirical Bayes flow matching (EB-FM), a method for learning flow matching generative priors from indirect observations by iteratively inverting a possibly non-linear forward model during training.
\item On controlled benchmarks we validate our design choices that stabilize the E-step and reduce the posterior sampling variance: flow matching with gradient based guidance and persistent latents, which reduce variance across iterations.
\item We apply EB-FM on simulated cryo-EM data and demonstrate the recovery of 3D conformational distributions from 2D projections directly in atomic-coordinate space, under known poses.
\end{enumerate}
Code is available at: \url{https://daiyral.github.io/Empirical-Bayes-Flow-Matching-for-Continuous-Cryo-EM-Heterogeneity/}

\section{Preliminaries}

\subsection{Flow Matching}
Continuous-time generative models learn to transport a simple, tractable base distribution $p_0$ (e.g., a standard normal) to a complex data distribution $p_1$ via an ordinary differential equation (ODE). Let $x_t \in \mathbb{R}^D$ denote the state of the system at time $t \in [0, 1]$. The Probability Flow ODE is defined by a time-dependent vector field $v_t$ formulated as: $dx_t = v_t(x_t) dt$.

Flow Matching \cite{liuFlow2023,lipman2023flowmatchinggenerativemodeling,albergo2025stochastic} provides a scalable framework to learn this vector field by regressing a neural network $v_\theta(x_t, t)$ against  conditional vector fields $v_t(x_t \mid x_1)$, defined given a data point $x_1 \sim p_1(x_1)$. The model parameters $\theta$ are optimized using the Conditional Flow Matching (CFM) objective:
\begin{equation}
\mathcal{L}_{\text{CFM}}(\theta) = \mathbb{E}_{t, x_1, x_t} \left[ \| v_\theta(x_t, t) - v_t(x_t \mid x_1) \|^2 \right]
\end{equation}
Evaluating this objective requires drawing clean samples $x_1$ from the target data distribution $p_1$. In the context of \textit{de novo} discovery of protein structure, clean 3D conformations are not given directly, but through a forward operator $y=\mathcal{A}(x)$, rendering the standard CFM objective intractable.

Given a generative prior $p_\theta(x)$ trained on clean data, sampling from the posterior $p_\theta(x \mid y) \propto p(y \mid x)\,p_\theta(x)$ requires augmenting the prior's reverse dynamics with the likelihood gradient $\nabla_{x_t} \log p(y \mid x_t)$, which has no closed form when the forward operator is non-linear. Diffusion Posterior Sampling (DPS) \cite{chungDiffusion2025} approximates this with a point-estimate of the network's $x$-prediction $\hat{x}_1(x_t, t)$, yielding a guidance term proportional to $\nabla_{x_t} \|y - \mathcal{A}(\hat{x}_1(x_t, t))\|^2$ that is back-propagated through the (possibly non-linear) forward operator $\mathcal{A}$. 
The same recipe extends naturally to flow matching \cite{kimFlowdps2025, pokleTraining2023}, and is the basis for our guided E-step in Section~\ref{sec:EBFM.Estep}.
Score-based priors were first applied to inverse problems in medical imaging~\cite{songSolving2023}, and posterior inference has since also been formulated variationally as stochastic
optimization over a parametric posterior~\cite{mardaniVariational2023}, an alternative to the sampling-based route we adopt.

For linear forward models, exact guidance becomes tractable via moment matching, and DiEM \cite{rozet2024learning} uses this to construct an empirical-Bayes method via a Conjugate-Gradient solver at every integration step. In non-linear settings with high dimensional observations such as in cryo-EM, exact moment matching is intractable, motivating DPS-style point-estimate guidance.
  
\subsection{The Cryo-EM Forward Model}
In cryo-EM, the observations are a set of noisy 2D projection images $\{y_i\}_{i=1}^N$. Let $x \in \mathbb{R}^{3N_a}$ denote the continuous 3D structure parametrized by the spatial coordinates of $N_a$ atoms. We assume the electron potential surrounding each atom can be approximated by a mixture of Gaussian radial basis functions. 
Under the simplifying assumption of known poses $R_i \in \text{SO}(3)$, the formation of an image $y_i$ evaluates the electron potential of the rotated atoms $R_i x$. Specifically, the pixel intensity $y_i(p)$ at a 2D coordinate $p \in \mathbb{R}^2$ is the line integral of this potential along the viewing axis $z$:
\begin{equation}
y_i(p) = \int \sum_{j=1}^{N_a} \sum_{k=1}^K w_{j,k} \mathcal{N}\left( \begin{bmatrix} p \\ z \end{bmatrix} \mathrel{;} R_i x_j, \sigma_{j,k}^2 I \right) dz + \epsilon_i
\end{equation}
where $\epsilon_i \sim \mathcal{N}(0, \sigma^2 I)$ models detector noise, and $\{w_{j,k}, \sigma_{j,k}\}$ parameterize the Gaussian mixture components (typically $K=5$) for the atomic potentials. 
This coordinate-based forward model defines a structurally non-linear rendering operator,  $y_i = \mathcal{A}_i(x) + \epsilon_i$, and  consequently a  highly non-linear likelihood function $p(y_i \mid x)$.

\subsection{Expectation-Maximization}
When the true 3D structures $\{x_i\}$ are unobserved, they must be treated as hidden variables. To learn the flow matching generative prior $p_\theta(x)$ directly from the observations $\{y_i\}$, one seeks to maximize the marginal log-likelihood:
\begin{equation}
\log p_\theta(y_i) = \log \int p(y_i \mid x_i) p_\theta(x_i) dx_i
\end{equation}
The standard Expectation-Maximization (EM) algorithm approaches this by alternating between two steps. The E-step computes the expected log-likelihood under the posterior distribution given the current parameters $\theta_{\text{old}}$, using
$
Q(\theta | \theta_{\text{old}}) = \mathbb{E}_{p_{\theta_{\text{old}}}(x_i \mid y_i)} [\log p(y_i \mid x_i) + \log p_\theta(x_i)]
$.
The M-step then updates the parameters by maximizing this auxiliary function: $\theta_{\text{new}} = \arg\max_\theta Q(\theta | \theta_{\text{old}})$. 


\section{Empirical Bayes Flow Matching}
\label{sec:EBFM}

We consider the general inverse problem of recovering an unknown continuous state $x \in \mathcal{X}$ from a noisy observation $y \in \mathcal{Y}$. 
The physical measurement process is governed by a known, potentially non-linear forward, or \emph{rendering}, operator $\mathcal{A}: \mathcal{X} \to \mathcal{Y}$ (e.g. the cryo-EM image acquisition process).
Assuming additive Gaussian noise $\epsilon \sim \mathcal{N}(0, \sigma^2_yI)$, the observation likelihood is given by:
\begin{equation}
p(y \mid x) = \mathcal{N}(y \mid \mathcal{A}(x), \sigma^2_y I)
\end{equation}
In the \textit{de novo} setting, we are given a dataset of $N$ independent observations $\mathcal{D} = \{y_i\}_{i=1}^N$, but no clean samples of the underlying states $x_i$. Our objective is to learn a flow matching generative prior $p_\theta(x)$ that accurately models the distribution of unobserved states, relying exclusively on the corrupted indirect measurements.

Because $x$ is unobserved, this naturally frames an Expectation-Maximization (EM) objective. The standard EM algorithm alternates between computing the posterior distribution $p_{\theta_{\text{old}}}(x \mid y_i)$ in the E-step, and maximizing the expected complete-data log-likelihood in the M-step. 
However, standard EM is  intractable in our setting, since $p_\theta(x)$ is only defined implicitly via the probability flow ODE of a continuous-time vector field $v_\theta(x_t, t)$, and the forward operator $\mathcal{A}(x)$ is non-linear. This prevents the analytical solution to the E step.  

\subsection{Stochastic Approximation of the Posterior Mean (E-Step)} \label{sec:EBFM.Estep}

To bypass the intractable E-step, we adopt a Monte Carlo Expectation-Maximization framework. Standard MCEM requires drawing a massive number of samples per observation to accurately compute the expected log-likelihood. This is computationally prohibitive for flow matching models that require multi-step sampling guided by back-propagating through the forward operator.  Conversely, using only a single posterior sample introduces severe variance, stemming also from the approximate nature of the posterior sampling trajectory itself.

We resolve this by explicitly tracking the posterior mean of each observation. 
We maintain a set of persistent latents $\{\tilde{x}_i\}_{i=1}^N$ corresponding to each observation $y_i$. In the E-step of iteration $k$, we first draw a single, noisy Monte Carlo sample $\hat{x}_i^{(k)}$ from the current posterior $p_{\theta_{k-1}}(x \mid y_i)$ using an approximate guided sampling procedure,  based on a point estimate of the predicted clean state, i.e. $\nabla_{x_t} \|\mathbf{y} - \mathcal{A}(\hat{x}_1(x_t, t))\|$ at each sampling time $t$. 
Then, rather than using this high-variance sample directly, we update our persistent latent using a stochastic approximation step:
\begin{equation} \label{eq:pl_ema}
\tilde{x}_i^{(k)} = (1 - \gamma_k) \tilde{x}_i^{(k-1)} + \gamma_k \hat{x}_i^{(k)}
\end{equation}
where $\gamma_k$ is the step size. In some experiments we also update the persistent latents with the rendering loss directly $\nabla_{\tilde{x}_i} \|\mathbf{y} - \mathcal{A}(\tilde{x}_i)\|$. This moving average approximately tracks the posterior mean, $\mathbb{E}[x \mid y_i]$, reflecting the assumption of local unimodality. In the cryo-EM setting, this means that while the global distribution can be highly complex, the posterior distribution of a conformation given a single specific 2D projection is well approximated by the posterior mean. 

\subsection{Flow Matching from Inferred States (M-Step)}
\label{sec:EBFM.Mstep}

The M-step trains the flow matching model on the persistent posterior samples $\{\tilde{x}_i^{(k)}\}_{i=1}^N$ in place of  the unobserved clean data. The Conditional Flow Matching (CFM) loss on a minibatch $\mathcal{B}$ becomes:
\begin{equation}
\mathcal{L}_{\text{CFM}}(\theta) = \frac{1}{|\mathcal{B}|} \sum_{i \in \mathcal{B}} \mathbb{E}_{t, x_0} \left[ \| v_\theta(x_t, t) - v_t(x_t \mid \tilde{x}_i^{(k)}) \|^2 \right]
\end{equation}
where $x_t = (1-t) x_0 + t \tilde{x}_i^{(k)} $, using $x_0 \sim p_0(x_0)$, $t \sim \mathcal{U}[0, 1]$ and $v_t(x_t \mid \tilde{x}_i^{(k)}) = \tilde{x}_i^{(k)} - x_0$. 


Each outer iteration runs one E-step epoch (updating all latents) followed by an M-step epoch (updating the model weights). See Alg.~\ref{alg:ebfm_general}. The full method is illustrated for cryo-EM in Figure~\ref{fig:overview}. 

\subsection{Controlled benchmarks}
\begin{figure}
\centering
\scriptsize{Linear: $\mathcal{A}(x_i) = A_ix_i$ \hspace{3cm} Non-linear: $\mathcal{A}(x_i) = sin(A_ix_i)$} \\
\includegraphics[width=1.0\textwidth,trim={0 0 0 1cm},clip]{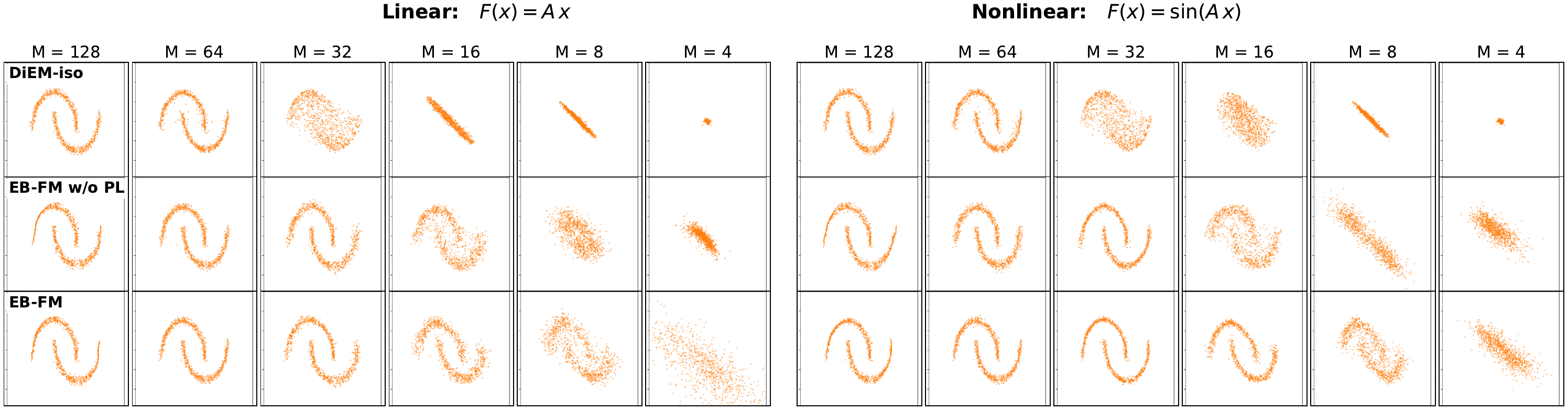}
\caption{Two-moons benchmark for linear and non-linear forward operators, across decreasing measurement dimensions $M$. Each panel shows samples from the learned prior projected back to 2D. DiEM-iso (top) is sharp at high $M$ but its linearized CG solver collapses by $M{=}32$ under increasing uncertainty. EB-FM without persistent latents (middle) preserves the manifold further into the under-determined regime but loses structure by $M{=}8$. EB-FM with persistent latents (bottom) is the only method that recovers the two moons at $M{=}8$.}
\label{fig:two_moons_scatter}
\end{figure}

To isolate the effects of our design choices, we first evaluate EB-FM on a controlled benchmark using the two-moons dataset. We sample 2D points $z_i \in \mathbb{R}^2$ and lift them into a high-dimensional space $x_i \in \mathbb{R}^D$ ($D=128$) via a fixed random matrix $P$. For each data sample, we draw a fixed random measurement matrix $A_i \in \mathbb{R}^{M \times D}$ and generate noisy observations:
$y_i = \mathcal{A}(x_i) + \varepsilon_i$, where $\varepsilon_i \sim \mathcal{N}(0,\sigma^2 I_M)$.
We test two forward operators: a linear model $\mathcal{A}(x_i) = A_i x_i$ and a non-linear model $\mathcal{A}(x_i) = \sin(A_i x_i)$. To evaluate performance across varying degrees of problem difficulty, we sweep the measurement dimension $M \in \{128, 64, 32, 16, 8, 4\}$, representing a transition from a fully-determined system to a highly under-determined regime ($M/D = 0.03$). 

We compare three methods, all sharing an identical multi-layer perceptron (MLP) architecture for the generative prior and an identical outer Expectation-Maximization loop structure:
(1) \textbf{DiEM-iso}: a tractable version of DiEM~\cite{rozet2024learning}, using a diffusion prior and a conjugate gradient (CG) solver on the linearized forward model, bypassing the need to back-propagate through the network at each CG step by assuming isotropic covariance of the prior. We implement this using the same scheduling as our method, running one E-step epoch and one M-step epoch at each outer iteration, as we find it performs better.  
(2) \textbf{EB-FM w/o PL}: Our flow matching method without persistent latents, i.e., latents are fully replaced at each outer EM iteration.
(3) \textbf{EB-FM}: Our full proposed method, updating the latents via an exponential moving average (EMA) across outer iterations. For more details about the implementations see App.~\ref{app:twomoons_details}.

\textbf{Results.} As shown in Figure~\ref{fig:two_moons_scatter}, all methods successfully recover the distribution in the fully-determined regime ($M=128$). However, as $M$ decreases and the posterior uncertainty grows, DiEM-iso's performance degrades, leading to gradual loss of structure at $M=32$ and mode collapse at smaller $M$.
Conversely, our gradient-guided EB-FM framework survives well into the under-determined regime, however without persistent latents (EB-FM w/o PL), the structure is lost at $M=8$.
With the addition of persistent latents, our model (EB-FM) successfully retains the underlying two-moons manifold geometry further into the under-determined regime, becoming the only method capable of resolving non-collapsed distributions at $M=8$.


\begin{figure}
\centering
\small{M=100 \hspace{150pt} M=50} \\ 
\includegraphics[width=0.48\linewidth,trim={0 0 0 2cm},clip]{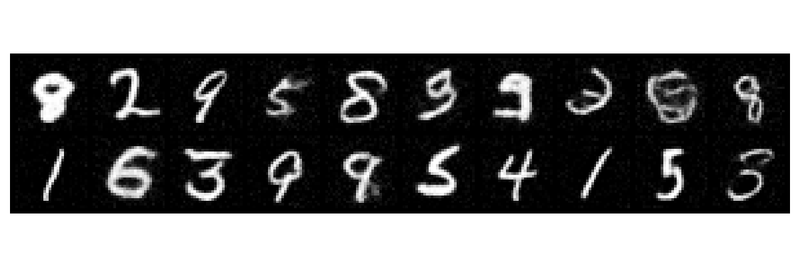}
\includegraphics[width=0.48\linewidth,trim={0 0 0 2cm},clip]{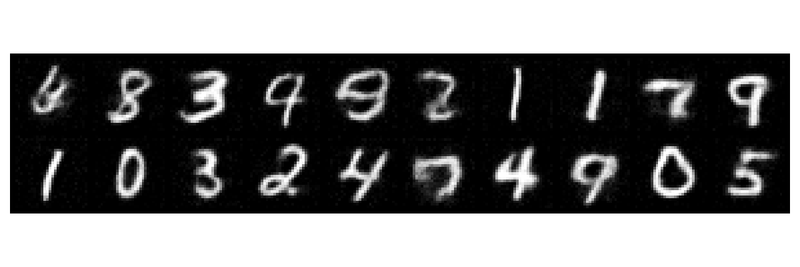}
\caption{{Unconditional samples from the EB-FM prior. 784-dimensional MNIST observed only through random Gaussian projections. Left: $M{=}100$ measurements per image. Right: $M{=}50$. The model never sees clean digits during training, yet recovers digit structure from the indirect observations.}}
\label{fig:mnist_samples}
\end{figure}

In addition to the two-moons data, we also test our method on MNIST data where the observation is a random projection from $D=784$ pixels to $M=100$ and $M=50$ observations.
Figure~\ref{fig:mnist_samples} shows resulting samples from the learned EB-FM prior, demonstrating that it captures the digit structure of the data even when during training the model has only access to low dimensional observations.
Comparisons with baselines and more details are provided in App.~\ref{app:mnist_details}.

\section{Cryo-EM heterogeneity in atomic-coordinate space}

We now apply EB-FM to the motivating problem of recovering the distribution of 3D protein conformations from noisy 2D cryo-EM projections, directly in atomic-coordinate space. 
We describe our experimental pipeline and the results on four generated datasets, one per protein, with EB-FM trained independently on each.
More information about the data and method implementation is provided in App.~\ref{app:cryoEM}.

\textbf{Datasets and Simulation}                     
  We evaluate our method on four proteins from the \textbf{ATLAS} database \cite{vander2024atlas}: \texttt{1a62\_A} (130 amino acids), \texttt{1egw\_C} (79 aa), \texttt{3rjs\_A} (89 aa), and \texttt{7r79\_C} (78  
  aa). 
  We use each protein to generate a separate dataset to train our model.
  These proteins are small to medium-sized structures commonly used in molecular dynamics studies and exhibit varying degrees of conformational flexibility. For all experiments, we use
  \emph{backbone atoms only} to reduce complexity while preserving the core structural signal. Each protein's MD ensemble contains $10{,}000$ conformations.      
  
  To generate observations, we simulate the cryo-electron microscopy (cryo-EM) image formation process using a high-quality TEM simulator~\cite{rullgaard2011simulation}. The simulator includes realistic physics  
  such as the contrast transfer function (CTF), detector effects, solvation, and noise models. For each protein conformation, we generate a single projection image by applying a random
  rotation as seen in Fig.~\ref{fig:cryo-em-samples}. The simulation follows standard practice, with parameters summarized in Table~\ref{tab:cryoem_params}.                           
  
  \textbf{Model}                                   
  The generative prior $p_\theta(x)$ is a Transformer operating on backbone atoms, where the 3D coordinate of each atom is treated as a token.
  The forward operator within our model is simpler than the simulator used to generate the data, and is implemented with CryoJAX~\cite{obrien2026cryojax}, a differentiable cryo-EM renderer in JAX, using a fixed single Gaussian per atom. 
     
\textbf{Training}             
Latents are initialized from the first trajectory frame of the corresponding Protein Data Bank (PDB) protein. When such a structure is unavailable, an initialization could be obtained using methods such as AlphaFold.
The forward loss used for posterior sampling and latent optimization is \(\mathcal{L}_{\mathrm{forward}}=\alpha\mathcal{L}_{\mathrm{render}}+\beta\mathcal{L}_{\mathrm{backbone}}\), where \(\mathcal{L}_{\mathrm{render}}=\|\mathcal{A}(\tilde{x}_i)-y_i\|_1\) compares the observed image \(y_i\) to the image produced by our differentiable cryo-EM renderer \(\mathcal{A}\) from the current latent structure, and \(\mathcal{L}_{\mathrm{backbone}}\) enforces structural consistency by penalizing squared deviations from pairwise inter-atomic distances of the initialization structure.
  
We begin with a \emph{warmup phase} that optimizes each $\tilde{x}_i$ by gradient descent against $ \mathcal{L_\text{forward}}$.
After warmup we run the EB-FM EM loop. In each E-step, for every observation we generate $5$ candidates with $3$ unguided integration steps, select the candidate with the lowest $\mathcal{L}_{\text{render}}$, and apply $10$ guided refinement steps using both rendering and backbone losses to produce a single posterior sample. 

We then update $\tilde{x}_i$ with one Adam step on the squared distance to the new
posterior sample plus $\mathcal{L}_{\text{forward}}$, the auxiliary losses used for
guidance. With a plain gradient step and without $\mathcal{L}_{\text{forward}}$, this
reduces to the EMA update of Eq \ref{eq:pl_ema}.

The M-step updates the flow model on the current $\{\tilde{x}_i\}$. We pre-train the flow for 50 epochs ($\approx2K$ updates), and then we alternate one E-step over all observations with one epoch of M-step updates per outer iteration.
Following \cite{jing2024alphafoldmeetsflowmatching} we use a \emph{harmonic prior} as the base distribution $p_0$, and align it to the latent target $\tilde{x}_i$ based on the root mean squared deviation (RMSD).

  \textbf{Evaluation}
    To evaluate EB-FM in the cryo-EM setting, we compare the predicted ensemble of protein conformations with the ground truth molecular dynamics (MD) ensemble using metrics that capture both flexibility and distributional accuracy. To assess flexibility, we measure the average pairwise root mean squared deviation (RMSD) within each ensemble, which reflects the overall structural variability, and compare it to the MD simulation. We further compute the root mean square fluctuation (RMSF) per residue, which quantifies local atomic deviations across the ensemble, and compare it to the ground truth, along with the Pearson correlation between the two RMSF profiles to evaluate whether the model correctly identifies regions of relative flexibility and rigidity along the protein. Beyond flexibility, we assess distributional accuracy by comparing the two ensembles as distributions rather than individual structures. Specifically, we approximate the Earth Mover’s Distance using RMSD as the ground metric (EMD-RMSD)~\cite{rosenbaum2021inferringcontinuousdistributionatom}: we sample $N=10{,}000$ structures from each ensemble and compute an $N \times N$ matrix of pairwise RMSDs after rigid alignment. We then solve an optimal one-to-one assignment using the Hungarian algorithm, and report the average RMSD over the matched pairs as the final EMD-RMSD score. Finally, we report coverage and fidelity in terms of RMSD. Coverage RMSD computes, for each ground truth structure, the minimum RMSD to any predicted sample, evaluating how well the predicted ensemble spans the true conformational space, while fidelity RMSD computes, for each predicted structure, the minimum RMSD to the ground truth, reflecting the accuracy and realism of the generated conformations.
            
  \subsection{Results}
\begin{figure}[t]
\centering
\includegraphics[width=\linewidth,height=0.55\textheight,keepaspectratio]{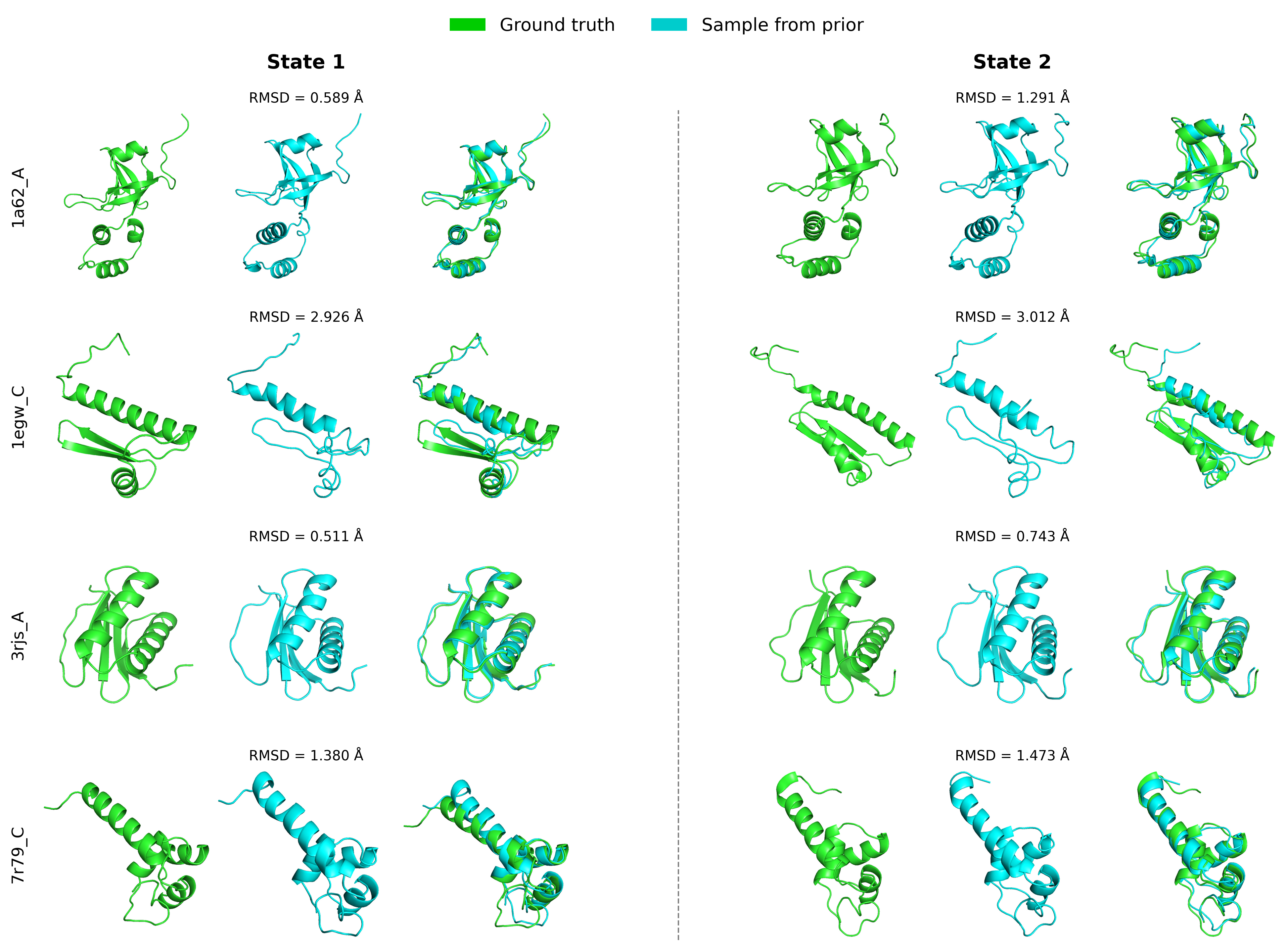}
\caption{Unconditional EB-FM samples on four ATLAS proteins, compared to ground-truth MD conformations. For each protein we show two ground-truth states selected to span the conformational change in the ensemble. Green: ground-truth. Cyan: the closest unconditional sample from the learned prior. Each pair is also shown superimposed with the corresponding RMSD. Animations of aligned trajectories are provided in the supplementary material.}
\label{fig:protein_samples}
\end{figure}

The samples shown in Figure~\ref{fig:protein_samples} demonstrate that EB-FM produces realistic and physically consistent conformations that closely match the ground truth structures across different states of each protein. The reported RMSD values are typically in the range of $0.51$--$1.30$~\AA\ for the more rigid proteins, such as 3rjs\_A and 1a62\_A, while the more flexible proteins, such as 7r79\_C and 1egw\_C, exhibit higher RMSD values reaching $3.0$~\AA. This reflects the increased difficulty of modeling broader conformational distributions from a single noisy 2D observation. The generated samples preserve the global fold and backbone geometry without introducing major artifacts, although small deviations are visible in flexible loop regions.
We provide animations of samples from the prior aligned to ground truth trajectories, in the supplementary material.

    \FloatBarrier
    \begin{figure}[t]
    \centering
    \includegraphics[width=0.43\linewidth]{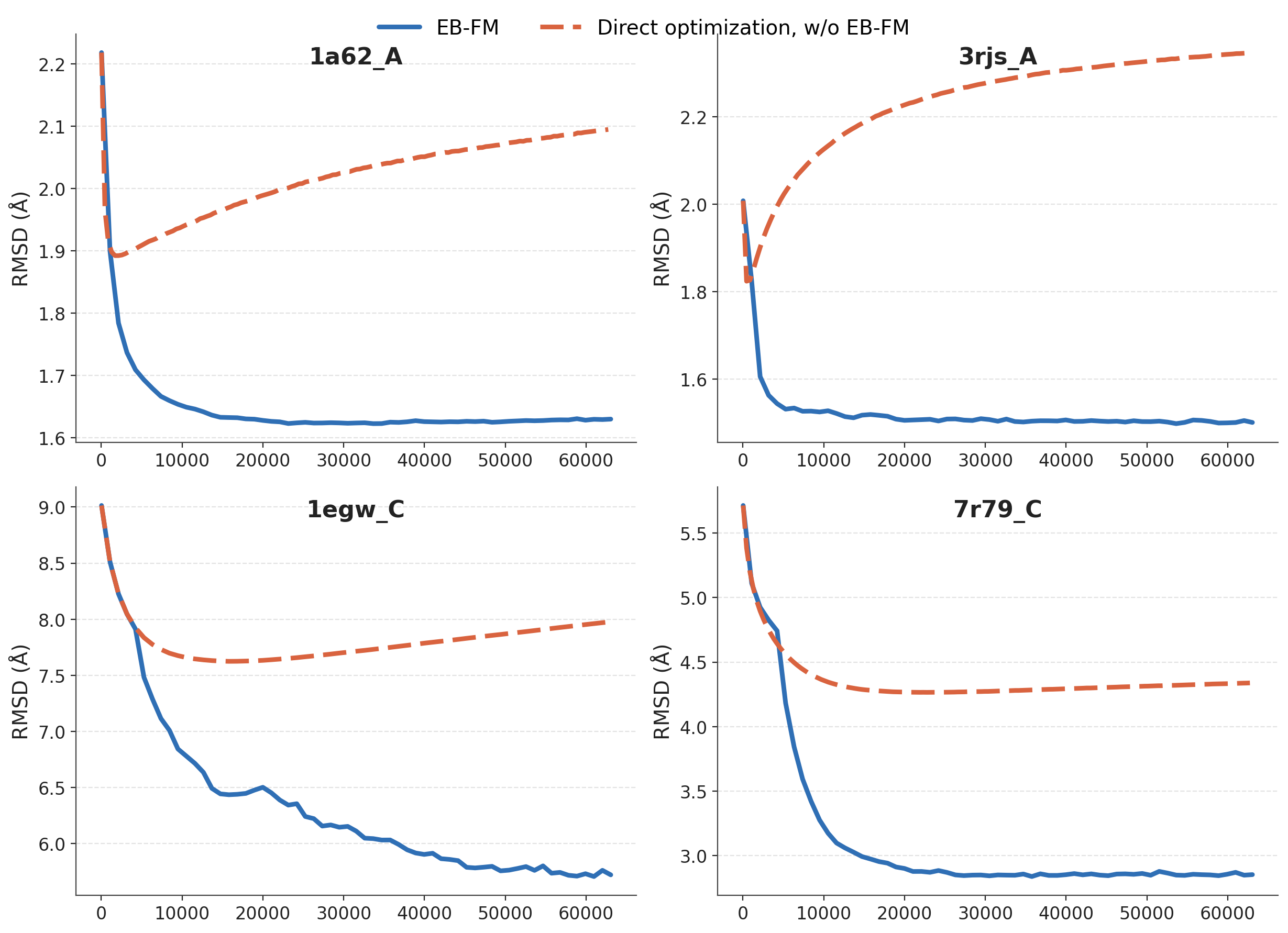}\hfill
    \includegraphics[width=0.57\linewidth]{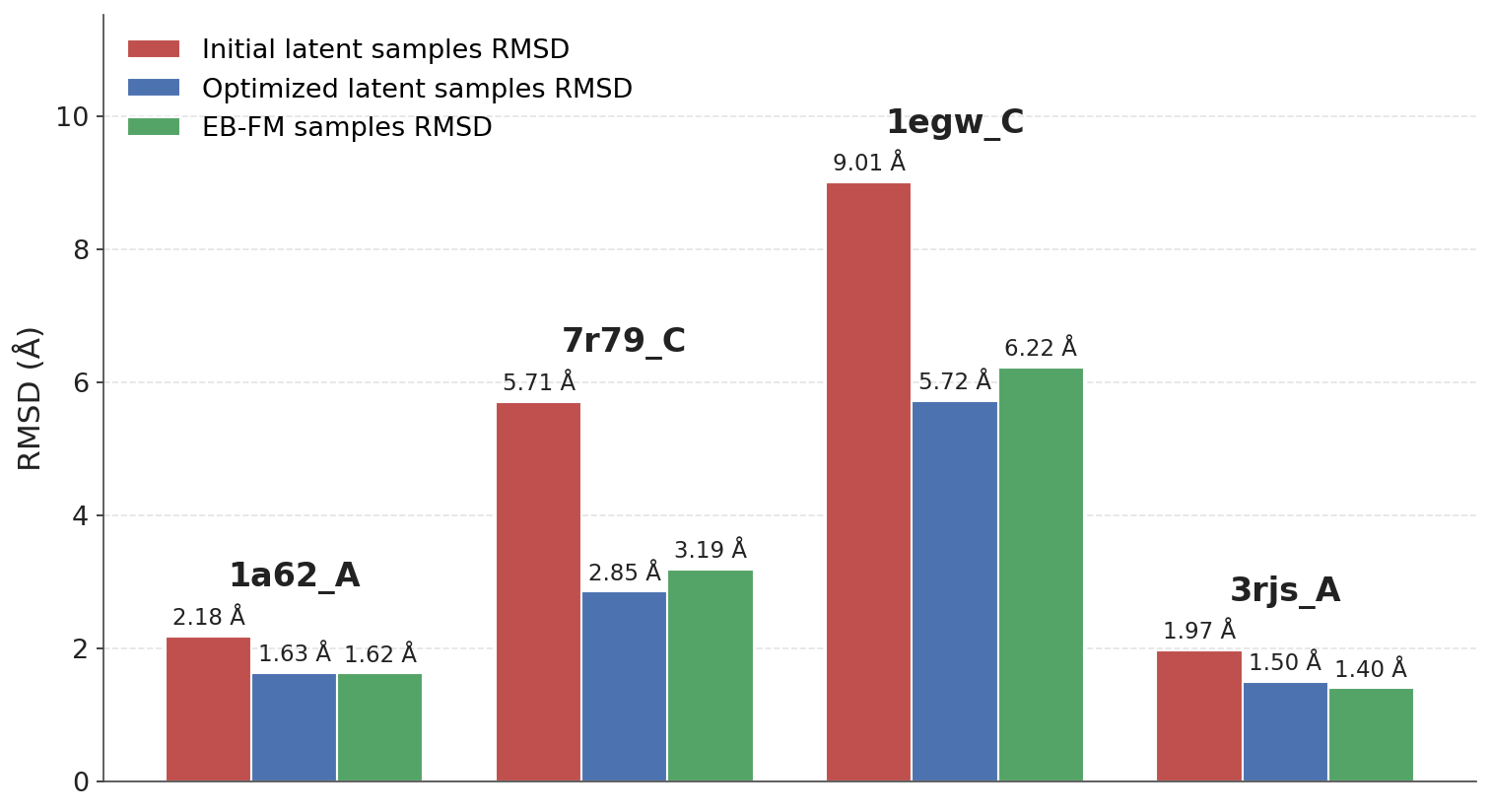}
    \caption{Left: Latent RMSD during optimization. EB-FM vs. direct latent optimization using only rendering and backbone losses (without learning flow matching prior).
    Right: Final RMSD comparison. Red: latent initialization using a reference structure; Blue: optimized latents after EB-FM training; Green: unconditional samples drawn from the learned flow matching prior.}
    \label{fig:rmsd_bar_plot}
    \end{figure}

    To further analyze reconstruction quality, Figure~\ref{fig:rmsd_bar_plot} (right) compares three quantities: the RMSD of the latent variables' initialization (which is a single representative structure), the RMSD of the final optimized latents after the full EB-FM procedure, and the RMSD of unconditional samples drawn directly from the trained flow matching prior. The optimized latents substantially improve over initialization across all proteins, indicating that the E-step posterior refinement successfully moves structures toward the true conformational ensemble. Importantly, the unconditional EB-FM samples achieve RMSD values close to those of the optimized latents, showing that the learned prior captures the underlying distribution. The significance of the empirical Bayes approach is also highlighted by the training dynamics in Figure~\ref{fig:rmsd_bar_plot} (left): Optimization based only on the rendering and backbone losses results in catastrophic divergence to degenerate structures (red curves), whereas the parallel training of the prior (EB-FM) facilitates the flow of information across observations gradually capturing the distribution of underlying conformations (blue curves).

    Table~\ref{tab:cryo_em_results} compares EB-FM with three complementary approaches for generating protein conformations. We compare against the VAE implemented by Rosenbaum et al \cite{rosenbaum2021inferringcontinuousdistributionatom}, which learns an autoencoder-based latent representation of protein structures and generates conformations from the learned latent space. We also compare against AlphaFold3 with MSA subsampling (AF3 MSA sub), which generates different conformations by randomly subsampling the multiple sequence alignment before each prediction \cite{del2022sampling, stein2022speach_af}. Finally, we compare against cryoDRGN \cite{zhong2021cryodrgn} combined with guided-AlphaFold3 (cryoDRGN-guided-AF3). In this approach, cryoDRGN reconstructs heterogeneous 3D density maps from the cryo-EM particle images, which are then used to guide AlphaFold3 during inference following the method of Maddipatla et al.\cite{maddipatla2026experiment}. Thus, while both this approach and EB-FM use the cryo-EM observations to recover conformational heterogeneity, EB-FM directly learns the distribution over atomic coordinates from the observations rather than first reconstructing density maps and subsequently guiding a pretrained structure predictor.

\begin{table*}[!t]
\vspace{-3pt}
\centering
\caption[Evaluation of generated conformational distributions across four proteins]{Evaluation of generated conformational distributions across four proteins. We compare EB-FM with the VAE implemented by Rosenbaum et al \cite{rosenbaum2021inferringcontinuousdistributionatom}, AlphaFold3 with MSA subsampling (AF3 MSA sub), and cryoDRGN combined with guided-AlphaFold3 (cryoDRGN-guided-AF3). \textbf{Distributional accuracy}: EMD-RMSD, Coverage RMSD, and Fidelity RMSD. \textbf{Flexibility}: pairwise RMSD, per-residue RMSF, and Pearson correlation $r$ between predicted and ground-truth RMSF profiles. Ground-truth (GT) values are shown for the flexibility metrics. Lower is better for the distributional accuracy metrics, and higher is better for RMSF $r$. For pairwise RMSD and RMSF, values closer to GT are better ($\rightarrow$ GT).\\}
\label{tab:cryo_em_results}
\vspace{-2pt}
\small
\setlength{\tabcolsep}{5pt}
\begin{tabular}{llccccc}
\toprule
\textbf{Protein} & \textbf{Metric} & \textbf{GT} & \textbf{EB-FM} & \textbf{VAE} & \textbf{AF3 MSA} & \textbf{cD+gAF3} \\
\midrule

\multirow{6}{*}{\textbf{1a62\_A}}
& EMD RMSD $\downarrow$
& -- & \textbf{1.62} & 1.80 & 1.83 & 1.80 \\
& Coverage RMSD $\downarrow$
& -- & \textbf{1.37} & 1.65 & 1.56 & 1.58 \\
& Fidelity RMSD $\downarrow$
& -- & 1.27 & 1.46 & 1.18 & \textbf{1.01} \\
& Pairwise RMSD $\rightarrow$ GT
& 2.40 & \textbf{1.58} & 1.20 & 1.23 & 0.77 \\
& RMSF $\rightarrow$ GT
& 1.08 & \textbf{0.77} & 0.60 & 0.59 & 0.51 \\
& RMSF $r$ $\uparrow$
& -- & 0.96 & \textbf{0.97} & 0.57 & 0.69 \\
\midrule

\multirow{6}{*}{\textbf{7r79\_C}}
& EMD RMSD $\downarrow$
& -- & \textbf{3.19} & 3.96 & 5.19 & 4.76 \\
& Coverage RMSD $\downarrow$
& -- & \textbf{2.35} & 3.11 & 3.22 & 3.05 \\
& Fidelity RMSD $\downarrow$
& -- & 2.35 & \textbf{2.06} & 4.47 & 3.72 \\
& Pairwise RMSD $\rightarrow$ GT
& 3.21 & 1.48 & 2.00 & 4.73 & \textbf{3.46} \\
& RMSF $\rightarrow$ GT
& 1.89 & 0.91 & 1.31 & 3.15 & \textbf{2.35} \\
& RMSF $r$ $\uparrow$
& -- & \textbf{0.91} & 0.85 & 0.64 & 0.61 \\
\midrule

\multirow{6}{*}{\textbf{1egw\_C}}
& EMD RMSD $\downarrow$
& -- & 6.22 & \textbf{5.56} & 8.30 & 7.87 \\
& Coverage RMSD $\downarrow$
& -- & \textbf{4.44} & 5.17 & 6.62 & 6.68 \\
& Fidelity RMSD $\downarrow$
& -- & 4.19 & 4.72 & \textbf{2.34} & 3.02 \\
& Pairwise RMSD $\rightarrow$ GT
& 6.45 & \textbf{4.29} & 3.06 & 1.98 & 2.85 \\
& RMSF $\rightarrow$ GT
& 3.11 & \textbf{2.55} & 1.55 & 0.74 & 0.83 \\
& RMSF $r$ $\uparrow$
& -- & \textbf{0.93} & 0.92 & 0.88 & 0.70 \\
\midrule

\multirow{6}{*}{\textbf{3rjs\_A}}
& EMD RMSD $\downarrow$
& -- & \textbf{1.40} & 2.05 & 1.56 & 1.41 \\
& Coverage RMSD $\downarrow$
& -- & 1.22 & 1.83 & \textbf{1.11} & 1.16 \\
& Fidelity RMSD $\downarrow$
& -- & 0.99 & 1.59 & 1.02 & \textbf{0.93} \\
& Pairwise RMSD $\rightarrow$ GT
& 1.79 & 1.10 & 0.93 & \textbf{1.17} & 0.94 \\
& RMSF $\rightarrow$ GT
& 0.68 & \textbf{0.49} & \textbf{0.49} & 0.36 & 0.45 \\
& RMSF $r$ $\uparrow$
& -- & \textbf{0.95} & 0.89 & \textbf{0.95} & 0.90 \\
\bottomrule
\end{tabular}
\vspace{-2pt}
\end{table*}

     In terms of EMD-RMSD, EB-FM outperforms cryoDRGN-guided-AF3 on all four proteins, with nearly identical performance on 3rjs\_A. For Coverage RMSD, EB-FM performs better on three of the four proteins, while for Fidelity RMSD it performs better on one. Compared with the VAE, EB-FM also achieves better results on the majority of metrics across the four proteins, although the VAE performs better on several metrics, particularly for 7r79\_C, and achieves a lower EMD-RMSD for 1egw\_C. These results suggest that learning a flow matching prior directly from observations can provide an effective representation of complex conformational distributions.

    The flexibility metrics further highlight the advantages of EB-FM. Its RMSF Pearson correlation ranges from $0.91$ to $0.96$ across the four proteins, exceeding that of cryoDRGN-guided-AF3 in every case and indicating that EB-FM accurately identifies the relative locations of flexible regions. In terms of absolute flexibility, measured by pairwise RMSD and RMSF, EB-FM produces ensembles that are closer to the ground-truth magnitude than cryoDRGN-guided-AF3 on three of the four proteins. However, EB-FM still underestimates the absolute magnitude of conformational motion relative to the ground-truth MD ensembles across all four proteins, as reflected in lower pairwise RMSD and RMSF values. Thus, although EB-FM captures which regions are flexible, it does not fully recover the magnitude of motion in those regions.
    
    The comparison with AF3 MSA subsampling further illustrates the difference between generating diversity by perturbing sequence-derived information and learning conformational heterogeneity directly from experimental observations. Although AlphaFold3 benefits from extensive structural knowledge acquired during training, its generated ensembles may not fully capture the conformational distributions supported by the cryo-EM observations. In contrast, EB-FM learns the conformational distribution directly from the observations while accounting for observation uncertainty.

    Additional ablation experiments can be seen in Table \ref {tab:cryo_em_ablations} which evaluate the robustness of EB-FM to CTF and pose errors, randomized initialization of the first frame, and removal of the backbone loss.
    
    Across the four proteins, Fidelity RMSD reaches as low as $0.99$~\AA, while Coverage RMSD is typically in the $1.2$--$2.3$~\AA\ range for three proteins, indicating that the generated samples remain close to valid ground-truth conformations while covering part of the true ensemble. The main limitations appear for the most flexible protein, 1egw\_C, where the higher EMD-RMSD and larger coverage and fidelity errors reflect the difficulty of recovering a broad, highly variable distribution from single noisy projections. Taken together, these results show that EB-FM learns a meaningful atomic-coordinate prior that recovers accurate structures, preserves the main flexible regions of the ensemble, and demonstrates the feasibility of learning conformational distributions directly from cryo-EM observations.

\section{Discussion}
In this work we introduced Empirical Bayes Flow Matching (EB-FM), a method for training a flow matching prior from indirect noisy observations, designed to make empirical Bayes efficient enough to handle high dimensional observations and non-linear forward models.
The standard cryo-EM pipeline first reconstructs a 3D density from 2D images and only then fits an atomic model, with no opportunity to use atomic-coordinate priors at the most under-determined stage of the problem. We showed that (1) combining single-sample DPS-style guidance with persistent latents updated by stochastic approximation gives an efficient empirical Bayes scheme, and (2) this enables bypassing the two-stage cryo-EM pipeline and training a strong atomic-coordinate prior \textit{de novo} directly from 2D projections. 
On two-moons and MNIST we validate the role of our design choices, and on simulated cryo-EM the learned prior produces realistic backbone conformations and recovers the flexibility profile of the ensemble.  

\textbf{Limitations and future work:} 
Our cryo-EM experiments assume known poses, while in real data the pose is an unknown parameter of the forward model. Posterior sampling with unknown forward parameters has been studied in the diffusion prior setting~\cite{murataGibbsDDRMPartiallyCollapsed2023}. A natural next step would be to adopt these approaches within our E-step. 
A second challenge, visible in our diversity metrics, is that the predicted ensembles underestimate the magnitude of conformational change. 
We expect this to become more limiting with unknown poses, where the local posterior $p(x | y_i)$ is more complex and multimodal and cannot be well summarized by its mean. One possible solution is to maintain multiple persistent latents per observation.
Finally, the method still remains computationally expensive at the current scale. We believe that stronger GPUs and better parallelization will make this approach practical for larger cryo-EM datasets and full atomic structures that contain side-chains rather than backbone atoms only.
Another promising direction is to initialize the flow matching prior from a model pretrained on known structures, such as CryoBoltz~\cite{raghu2025cryoboltz}, combining a strong starting prior with the ability of EB-FM to refine it directly on the observed data.

\begin{ack}
We thank Rachel Kolodny for her guidance on protein structure and conformational heterogeneity, and for her help in choosing the data, proteins, evaluations and experiments. This research was supported by the NVIDIA Academic Grant Program, using NVIDIA H100 GPUs on NVIDIA Brev.
\end{ack}

\bibliographystyle{unsrt}
\bibliography{mybib}

@article{jumper2021highly,
  title={Highly accurate protein structure prediction with AlphaFold},
  author={Jumper, John and Evans, Richard and Pritzel, Alexander and Green, Tim and Figurnov, Michael and Ronneberger, Olaf and Tunyasuvunakool, Kathryn and Bates, Russ and {\v{Z}}{\'\i}dek, Augustin and Potapenko, Anna and others},
  journal={nature},
  volume={596},
  number={7873},
  pages={583--589},
  year={2021},
  publisher={Nature Publishing Group}
}

@article{nogales2015cryo,
  title={Cryo-EM: a unique tool for the visualization of macromolecular complexity},
  author={Nogales, Eva and Scheres, Sjors HW},
  journal={Molecular cell},
  volume={58},
  number={4},
  pages={677--689},
  year={2015},
  publisher={Elsevier}
}

@article{GIRI2023102536,
title = {Deep learning for reconstructing protein structures from cryo-EM density maps: Recent advances and future directions},
journal = {Current Opinion in Structural Biology},
volume = {79},
pages = {102536},
year = {2023},
issn = {0959-440X},
doi = {https://doi.org/10.1016/j.sbi.2023.102536},
url = {https://www.sciencedirect.com/science/article/pii/S0959440X23000106},
author = {Nabin Giri and Raj S. Roy and Jianlin Cheng}
}

@article{egelman2016current,
  title={The current revolution in cryo-EM},
  author={Egelman, Edward H},
  journal={Biophysical journal},
  volume={110},
  number={5},
  pages={1008--1012},
  year={2016},
  publisher={Elsevier}
}

@article{frank2002single,
  title={Single-particle imaging of macromolecules by cryo-electron microscopy},
  author={Frank, Joachim},
  journal={Annual review of biophysics and biomolecular structure},
  volume={31},
  number={1},
  pages={303--319},
  year={2002},
  publisher={Annual Reviews 4139 El Camino Way, PO Box 10139, Palo Alto, CA 94303-0139, USA}
}

@article{zhong2021cryodrgn,
  title={CryoDRGN: reconstruction of heterogeneous cryo-EM structures using neural networks},
  author={Zhong, Ellen D and Bepler, Tristan and Berger, Bonnie and Davis, Joseph H},
  journal={Nature methods},
  volume={18},
  number={2},
  pages={176--185},
  year={2021},
  publisher={Nature Publishing Group US New York}
}

@misc{ducrocq2025cryospheresingleparticleheterogeneousreconstruction,
      title={cryoSPHERE: Single-particle heterogeneous reconstruction from cryo EM}, 
      author={Gabriel Ducrocq and Lukas Grunewald and Sebastian Westenhoff and Fredrik Lindsten},
      year={2025},
      eprint={2407.01574},
      archivePrefix={arXiv},
      primaryClass={q-bio.BM},
      url={https://arxiv.org/abs/2407.01574}, 
}

@misc{rosenbaum2021inferringcontinuousdistributionatom,
      title={Inferring a Continuous Distribution of Atom Coordinates from Cryo-EM Images using VAEs}, 
      author={Dan Rosenbaum and Marta Garnelo and Michal Zielinski and Charlie Beattie and Ellen Clancy and Andrea Huber and Pushmeet Kohli and Andrew W. Senior and John Jumper and Carl Doersch and S. M. Ali Eslami and Olaf Ronneberger and Jonas Adler},
      year={2021},
      eprint={2106.14108},
      archivePrefix={arXiv},
      primaryClass={cs.CE},
      url={https://arxiv.org/abs/2106.14108}, 
}

@misc{levy2024mixtureneuralfieldsheterogeneous,
      title={Mixture of neural fields for heterogeneous reconstruction in cryo-EM}, 
      author={Axel Levy and Rishwanth Raghu and David Shustin and Adele Rui-Yang Peng and Huan Li and Oliver Biggs Clarke and Gordon Wetzstein and Ellen D. Zhong},
      year={2024},
      eprint={2412.09420},
      archivePrefix={arXiv},
      primaryClass={cs.LG},
      url={https://arxiv.org/abs/2412.09420}, 
}

@inproceedings{lipman2023flowmatchinggenerativemodeling,
      title={Flow Matching for Generative Modeling}, 
      author={Yaron Lipman and Ricky T. Q. Chen and Heli Ben-Hamu and Maximilian Nickel and Matt Le},
      year={2023},
      booktitle = {International Conference on Learning Representations (ICLR)},
      eprint={2210.02747},
      archivePrefix={arXiv},
      primaryClass={cs.LG},
      url={https://arxiv.org/abs/2210.02747}, 
}

@article{punjani2017cryosparc,
  title={cryoSPARC: algorithms for rapid unsupervised cryo-EM structure determination},
  author={Punjani, Ali and Rubinstein, John L and Fleet, David J and Brubaker, Marcus A},
  journal={Nature methods},
  volume={14},
  number={3},
  pages={290--296},
  year={2017},
  publisher={Nature Publishing Group}
}

@article{scheres2012relion,
  title={RELION: implementation of a Bayesian approach to cryo-EM structure determination},
  author={Scheres, Sjors HW},
  journal={Journal of structural biology},
  volume={180},
  number={3},
  pages={519--530},
  year={2012},
  publisher={Elsevier}
}

@misc{ho2020denoisingdiffusionprobabilisticmodels,
      title={Denoising Diffusion Probabilistic Models}, 
      author={Jonathan Ho and Ajay Jain and Pieter Abbeel},
      year={2020},
      eprint={2006.11239},
      archivePrefix={arXiv},
      primaryClass={cs.LG},
      url={https://arxiv.org/abs/2006.11239}, 
}

@article{jing2023eigenfold,
  title={Eigenfold: Generative protein structure prediction with diffusion models},
  author={Jing, Bowen and Erives, Ezra and Pao-Huang, Peter and Corso, Gabriele and Berger, Bonnie and Jaakkola, Tommi},
  journal={arXiv preprint arXiv:2304.02198},
  year={2023}
}

@article{vulovic2013image,
  title={Image formation modeling in cryo-electron microscopy},
  author={Vulovi{\'c}, Milo{\v{s}} and Ravelli, Raimond BG and van Vliet, Lucas J and Koster, Abraham J and Lazi{\'c}, Ivan and L{\"u}cken, Uwe and Rullg{\aa}rd, Hans and {\"O}ktem, Ozan and Rieger, Bernd},
  journal={Journal of structural biology},
  volume={183},
  number={1},
  pages={19--32},
  year={2013},
  publisher={Elsevier}
}

@misc{jing2024alphafoldmeetsflowmatching,
      title={AlphaFold Meets Flow Matching for Generating Protein Ensembles}, 
      author={Bowen Jing and Bonnie Berger and Tommi Jaakkola},
      year={2024},
      eprint={2402.04845},
      archivePrefix={arXiv},
      primaryClass={q-bio.BM},
      url={https://arxiv.org/abs/2402.04845}, 
}

@article{levy2024solving,
  title={Solving inverse problems in protein space using diffusion-based priors},
  author={Levy, Axel and Chan, Eric R and Fridovich-Keil, Sara and Poitevin, Fr{\'e}d{\'e}ric and Zhong, Ellen D and Wetzstein, Gordon},
  journal={arXiv preprint arXiv:2406.04239},
  year={2024}
}

@article{vander2024atlas,
  title={ATLAS: protein flexibility description from atomistic molecular dynamics simulations},
  author={Vander Meersche, Yann and Cretin, Gabriel and Gheeraert, Aria and Gelly, Jean-Christophe and Galochkina, Tatiana},
  journal={Nucleic acids research},
  volume={52},
  number={D1},
  pages={D384--D392},
  year={2024},
  publisher={Oxford University Press}
}

@article{rullgaard2011simulation,
  title={Simulation of transmission electron microscope images of biological specimens},
  author={Rullg{\aa}rd, Hans and {\"O}fverstedt, L-G and Masich, Sergey and Daneholt, Bertil and {\"O}ktem, Ozan},
  journal={Journal of microscopy},
  volume={243},
  number={3},
  pages={234--256},
  year={2011},
  publisher={Wiley Online Library}
}

@inproceedings{sohl2015deep,
  author    = {Sohl-Dickstein, Jascha and Weiss, Eric and Maheswaranathan, Niru and Ganguli, Surya},
  title     = {Deep unsupervised learning using nonequilibrium thermodynamics},
  booktitle = {Proceedings of the 32nd International Conference on Machine Learning (ICML)},
  series    = {PMLR},
  volume    = {37},
  pages     = {2256--2265},
  year      = {2015}
}

@inproceedings{song2021scorebased,
  author    = {Song, Yang and Sohl-Dickstein, Jascha and Kingma, Diederik P. and Kumar, Abhishek and Ermon, Stefano and Poole, Ben},
  title     = {Score-based generative modeling through stochastic differential equations},
  booktitle = {International Conference on Learning Representations (ICLR)},
  year      = {2021},
  note      = {Outstanding Paper Award}
}

@article{albergo2025stochastic,
  author  = {Albergo, Michael S. and Boffi, Nicholas M. and Vanden-Eijnden, Eric},
  title   = {Stochastic interpolants: A unifying framework for flows and diffusions},
  journal = {Journal of Machine Learning Research},
  volume  = {26},
  number  = {209},
  pages   = {1--80},
  year    = {2025}
}

@article{punjani20233dflex,
  author  = {Punjani, Ali and Fleet, David J.},
  title   = {{3DFlex}: determining structure and motion of flexible proteins from cryo-{EM}},
  journal = {Nature Methods},
  volume  = {20},
  pages   = {860--870},
  year    = {2023},
  doi     = {10.1038/s41592-023-01853-8}
}

@inproceedings{levy2022cryofire,
  author    = {Levy, Axel and Wetzstein, Gordon and Martel, Julien N. P. and Poitevin, Fr{\'e}d{\'e}ric and Zhong, Ellen D.},
  title     = {Amortized inference for heterogeneous reconstruction in cryo-{EM}},
  booktitle = {Advances in Neural Information Processing Systems (NeurIPS)},
  volume    = {35},
  year      = {2022}
}

@article{ingraham2023chroma,
  author  = {Ingraham, John B. and Baranov, Max and Costello, Zak and Barber, Karl W. and Wang, Wujie and Ismail, Ahmed and Frappier, Vincent and Lord, Dana M. and Ng-Thow-Hing, Christopher and Van Vlack, Erik R. and Tie, Shan and Xue, Vincent and Cowles, Sarah C. and Leung, Alan and Rodrigues, Jo{\~a}o V. and Morales-Perez, Claudio L. and Ayoub, Alex M. and Green, Robin and Puentes, Katherine and Oplinger, Frank and Panwar, Nishant V. and Obermeyer, Fritz and Root, Adam R. and Beam, Andrew L. and Poelwijk, Frank J. and Grigoryan, Gevorg},
  title   = {Illuminating protein space with a programmable generative model},
  journal = {Nature},
  volume  = {623},
  number  = {7989},
  pages   = {1070--1078},
  year    = {2023},
  doi     = {10.1038/s41586-023-06728-8}
}

@article{watson2023rfdiffusion,
  author  = {Watson, Joseph L. and Juergens, David and Bennett, Nathaniel R. and Trippe, Brian L. and Yim, Jason and Eisenach, Helen E. and Ahern, Woody and Borst, Andrew J. and Ragotte, Robert J. and Milles, Lukas F. and Wicky, Basile I. M. and Hanikel, Nikita and Pellock, Samuel J. and Courbet, Alexis and Sheffler, William and Wang, Jue and Venkatesh, Preetham and Sappington, Isaac and Torres, Susana V{\'a}zquez and Lauko, Anna and De Bortoli, Valentin and Mathieu, Emile and Ovchinnikov, Sergey and Barzilay, Regina and Jaakkola, Tommi S. and DiMaio, Frank and Baek, Minkyung and Baker, David},
  title   = {De novo design of protein structure and function with {RFdiffusion}},
  journal = {Nature},
  volume  = {620},
  number  = {7976},
  pages   = {1089--1100},
  year    = {2023},
  doi     = {10.1038/s41586-023-06415-8}
}

@inproceedings{rozet2024learning,
  author    = {Rozet, Fran{\c{c}}ois and Andry, G{\'e}r{\^o}me and Lanusse, Fran{\c{c}}ois and Louppe, Gilles},
  title     = {Learning diffusion priors from observations by expectation maximization},
  booktitle = {Advances in Neural Information Processing Systems (NeurIPS)},
  volume    = {37},
  year      = {2024}
}

@article{delyon1999convergence,
  author  = {Delyon, Bernard and Lavielle, Marc and Moulines, Eric},
  title   = {Convergence of a stochastic approximation version of the {EM} algorithm},
  journal = {The Annals of Statistics},
  volume  = {27},
  number  = {1},
  pages   = {94--128},
  year    = {1999},
  doi     = {10.1214/aos/1018031103}
}

@inproceedings{raghu2025cryoboltz,
  title = {Multiscale guidance of protein structure prediction with heterogeneous cryo-EM data},
  author = {Raghu, Rishwanth and Levy, Axel and Wetzstein, Gordon and Zhong, Ellen D.},
  booktitle = {Advances in Neural Information Processing Systems (NeurIPS)},
  year = {2025}
}

@article{frauenfelder1991energy,
  title={The energy landscapes and motions of proteins},
  author={Frauenfelder, Hans and Sligar, Stephen G and Wolynes, Peter G},
  journal={Science},
  volume={254},
  number={5038},
  pages={1598--1603},
  year={1991},
  publisher={American Association for the Advancement of Science}
}

@article{motlagh2014ensemble,
  title={The ensemble nature of allostery},
  author={Motlagh, Hesam N and Wrabl, James O and Li, Jing and Hilser, Vincent J},
  journal={Nature},
  volume={508},
  number={7496},
  pages={331--339},
  year={2014},
  publisher={Nature Publishing Group UK London}
}

@article{abramson2024accurate,
  title={Accurate structure prediction of biomolecular interactions with {AlphaFold} 3},
  author={Abramson, Josh and Adler, Jonas and Dunger, Jack and Evans, Richard and Green, Tim and Pritzel, Alexander and Ronneberger, Olaf and Willmore, Lindsay and Ballard, Andrew J and Bambrick, Joshua and others},
  journal={Nature},
  volume={630},
  number={8016},
  pages={493--500},
  year={2024},
  publisher={Nature Publishing Group UK London}
}

@article{obrien2026cryojax,
author = "O'Brien, Michael J. and Silva-S{\'{a}}nchez, David and Woollard, Geoffrey and Je, Kwanghwi and Hanson, Sonya M. and Needleman, Daniel J. and Cossio, Pilar and Thiede, Erik Henning and Astore, Miro A.",
title = "{CryoJAX: a cryo-electron microscopy image-simulation library in JAX}",
journal = "Acta Crystallographica Section D",
year = "2026",
volume = "82",
number = "3",
pages = "",
month = "Mar",
doi = {10.1107/S2059798326000550},
url = {https://doi.org/10.1107/S2059798326000550},
}

@inproceedings{chungDiffusion2025,                                   title     = {Diffusion Posterior Sampling for General Noisy Inverse Problems},
    author    = {Hyungjin Chung and Jeongsol Kim and Michael T. McCann and Marc L. Klasky and Jong Chul Ye},                 
    booktitle = {International Conference on Learning Representations (ICLR)},                                     
    year      = {2023},                                          
    eprint    = {2209.14687},                                    
    archiveprefix = {arXiv}                                      
  }

@inproceedings{songSolving2023,                                
    title     = {Solving Inverse Problems in Medical Imaging with Score-Based Generative Models},
    author    = {Yang Song and Liyue Shen and Lei Xing and Stefano Ermon},
    booktitle = {International Conference on Learning Representations (ICLR)},                                     
    year      = {2022},
    eprint    = {2111.08005},                                    
    archiveprefix = {arXiv}                                      
  }

@article{mardaniVariational2023,
    title     = {A Variational Perspective on Solving Inverse Problems with Diffusion Models},
    author    = {Morteza Mardani and Jiaming Song and Jan Kautz and Arash Vahdat},                                           
    journal   = {arXiv preprint arXiv:2305.04391},               
    year      = {2023},                                          
    eprint    = {2305.04391},                                    
    archiveprefix = {arXiv}                                      
  }

@inproceedings{kimFlowdps2025,                                     title     = {{FlowDPS}: Flow-Driven Posterior Sampling for Inverse Problems},
    author    = {Jeongsol Kim and Bryan Sangwoo Kim and Jong Chul Ye},                                                    
    booktitle = {IEEE/CVF International Conference on Computer Vision (ICCV)},                                              
    year      = {2025},                                          
    eprint    = {2503.08136},                                    
    archiveprefix = {arXiv}                                      
  }

@article{pokleTraining2023,
    title     = {Training-Free Linear Image Inverses via Flows},
    author    = {Ashwini Pokle and Matthew J. Muckley and Ricky T. Q. Chen and Brian Karrer}, 
    journal   = {arXiv preprint arXiv:2310.04432},  
    year      = {2023},                             
    eprint    = {2310.04432},                       archiveprefix = {arXiv}                         
  }

@inproceedings{liuFlow2023,
  title = {Flow Straight and Fast: Learning to Generate and Transfer Data with Rectified Flow},
  author = {Xingchao Liu and Chengyue Gong and Qiang Liu},
  year = {2023},
  booktitle = {International Conference on Learning Representations (ICLR)},
  doi = {10.48550/arXiv.2209.03003},
  eprint = {2209.03003},
  archiveprefix = {arXiv}
}

@inproceedings{darasAmbient2023,
  title = {Ambient Diffusion: {{Learning}} Clean Distributions from Corrupted Data},
  booktitle = {Advances in Neural Information Processing Systems (NeurIPS)},
  author = {Daras, Giannis and Shah, Kulin and Dagan, Yuval and Gollakota, Aravind and Dimakis, Alexandros G. and Klivans, Adam},
  year = {2023},
  eprint = {2305.19256},
  eprinttype = {arXiv},
}

@inproceedings{murataGibbsDDRMPartiallyCollapsed2023,
  title = {{{GibbsDDRM}}: {{A Partially Collapsed Gibbs Sampler}} for {{Solving Blind Inverse Problems}} with {{Denoising Diffusion Restoration}}},
  booktitle = {Int. {{Conf}}. on {{Machine Learning}}},
  author = {Murata, Naoki and Saito, Koichi and Lai, Chieh-Hsin and Takida, Yuhta and Uesaka, Toshimitsu and Mitsufuji, Yuki and Ermon, Stefano},
  date = {2023}
}

@article{casellaIntroductionEmpiricalBayes1985,
  title = {An {{Introduction}} to {{Empirical Bayes Data Analysis}}},
  author = {Casella, George},
  year = {1985},
  journaltitle = {The American Statistician},
  volume = {39},
  number = {2},
  eprint = {2682801},
  eprinttype = {jstor},
  pages = {83--87},
  publisher = {[American Statistical Association, Taylor \& Francis, Ltd.]},
  issn = {0003-1305},
  doi = {10.2307/2682801},
  url = {https://www.jstor.org/stable/2682801},
  urldate = {2026-05-07}
}

@article{efronTwoModelingStrategies2014,
  title = {Two Modeling Strategies for Empirical {{Bayes}} Estimation},
  author = {Efron, Bradley},
  date = {2014-05},
  year = {2014},
  journal = {Statistical science : a review journal of the Institute of Mathematical Statistics},
  shortjournal = {Stat Sci},
  volume = {29},
  number = {2},
  eprint = {25324592},
  eprinttype = {pubmed},
  pages = {285--301},
  issn = {0883-4237},
  doi = {10.1214/13-sts455},
  url = {https://pmc.ncbi.nlm.nih.gov/articles/PMC4196219/},
  urldate = {2026-05-07},
  pmcid = {PMC4196219}
}

@article{robbins1964empirical,
  title={The empirical Bayes approach to statistical decision problems},
  author={Robbins, Herbert},
  journal={The Annals of Mathematical Statistics},
  volume={35},
  number={1},
  pages={1--20},
  year={1964},
  publisher={JSTOR}
}

@article{maddipatla2026experiment,
  title={Experiment-guided AlphaFold3 resolves measurement-consistent protein ensembles},
  author={Maddipatla, Advaith and Sellam Bojan, Nadav and Bojan, Meital and Masalitin, Volodymyr and Vedula, Sanketh and Schanda, Paul and Marx, Ailie and Bronstein, Alex M},
  journal={Nature Biotechnology},
  pages={1--9},
  year={2026},
  publisher={Nature Publishing Group US New York}
}

@article{del2022sampling,
  title={Sampling alternative conformational states of transporters and receptors with AlphaFold2},
  author={Del Alamo, Diego and Sala, Davide and Mchaourab, Hassane S and Meiler, Jens},
  journal={elife},
  volume={11},
  pages={e75751},
  year={2022},
  publisher={eLife Sciences Publications, Ltd}
}

@article{stein2022speach_af,
  title={SPEACH\_AF: Sampling protein ensembles and conformational heterogeneity with Alphafold2},
  author={Stein, Richard A and Mchaourab, Hassane S},
  journal={PLoS computational biology},
  volume={18},
  number={8},
  pages={e1010483},
  year={2022},
  publisher={Public Library of Science San Francisco, CA USA}
}

\newpage
\appendix

\begin{algorithm}[t]
\caption{Empirical Bayes Flow Matching}
\label{alg:ebfm_general}
\begin{algorithmic}[1]
\Require Observations $\{y_i\}_{i=1}^N$, forward operator $\mathcal{A}$, flow model $f_\theta$
\Require Persistent latents $\{\tilde{x}_i\}_{i=1}^N$, base distribution $p_0$
\Require Outer iterations $K$, sampling steps $N_s$, flow-training steps $N_f$
\Require EMA step size $\gamma$, guidance strength $\eta$

\For{$k = 1$ to $K$}

    \Statex \textbf{E-step: guided posterior samples}
    \For{$i = 1$ to $N$}
        \State $x_t \sim p_0$
        \For{$n = 0$ to $N_s-1$}
            \State $t \gets n/N_s$, \quad $s \gets t+1/N_s$
            \State $\hat{x}_1 \gets f_\theta(x_t,t)$
            \State $\mathcal{L}_{\mathrm{render}} \gets \|\mathcal{A}(\hat{x}_1)-y_i\|_2^2$
            \State $v \gets (\hat{x}_1-x_t)/(1-t)$
            \State $x_t \gets x_t+(s-t)v-\eta \nabla_{x_t}\mathcal{L}_{\mathrm{render}}$
        \EndFor
        \State $\tilde{x}_i \gets (1-\gamma)\tilde{x}_i+\gamma x_t$
    \EndFor

    \Statex \textbf{M-step: train flow prior on latents}
    \For{$m = 1$ to $N_f$}
        \State Sample minibatch $B$
        \State $x_1 \gets \{\tilde{x}_i : i\in B\}$
        \State Sample $x_0 \sim p_0$, \quad $t\sim\mathcal{U}(0,1)$
        \State $x_t \gets (1-t)x_0+tx_1$
        \State $\hat{x}_1 \gets f_\theta(x_t,t)$
        \State $\mathcal{L}_{\mathrm{FM}} \gets
        \left\|
        \dfrac{\hat{x}_1-x_t}{1-t}
        -
        \dfrac{x_1-x_t}{1-t}
        \right\|_2^2$
        \State Update $\theta$ using $\nabla_\theta\mathcal{L}_{\mathrm{FM}}$
    \EndFor

\EndFor

\State \textbf{return} $f_\theta$, $\{\tilde{x}_i\}_{i=1}^N$
\end{algorithmic}
\end{algorithm}

\section{Two-Moons Experimental Details}
\label{app:twomoons_details}

\paragraph{Dataset and problem setup.}
We use the two-moons distribution using the function  \texttt{sklearn.datasets.make\_moons}
($n_\text{train}=10{,}000$, noise $0.05$), standardized to zero mean and
unit variance per coordinate. Each 2D sample $z_i\in\mathbb{R}^2$ is lifted
to $x_i = P z_i \in \mathbb{R}^D$ using a fixed column-orthonormal
embedding $P\in\mathbb{R}^{D\times 2}$ drawn once as the orthogonal
factor of a QR decomposition of an i.i.d.\ Gaussian matrix; we use
$D=128$. For each sample we draw an independent measurement matrix
$A_i\in\mathbb{R}^{M\times D}$ with i.i.d.\ $\mathcal{N}(0,1/D)$ entries,
fixed for the lifetime of $i$, and form the observation
$y_i=\mathcal{A}(x_i)+\varepsilon_i$ with
$\varepsilon_i\sim\mathcal{N}(0,\sigma_y^2 I_M)$ and $\sigma_y=10^{-3}$.
We evaluate two forward operators, $\mathcal{A}(x)=A_i x$ and
$\mathcal{A}(x)=\sin(A_i x)$, and sweep
$M\in\{128,64,32,16,8,4\}$.

\paragraph{Flow model architecture.}
All three methods share the same $x$-prediction multilayer perceptron:
5 hidden ReLU layers of width 256, with a sinusoidal time embedding
(dim 128) projected and concatenated to $x_t$. EB-FM and EB-FM w/o PL
train this network with the flow matching $v$-loss; DiEM-iso trains
it with the analogous $v$-loss for VP-DDPM under a linear $\beta$
schedule with $T=1000$.

\paragraph{EB-FM guided sampling.}
The E-step uses explicit Euler integration with $N=20$ steps. At each
step we predict $\hat{x}_1=f_\theta(x_t,t)$, compute the
rendering-loss gradient $g=\nabla_{x_t}\|A_i\hat{x}_1-y_i\|_2^2$ via
back-propagation through the network, and combine the prior velocity
and gradient correction in a single update,
\[
x_{t+\Delta t} = x_t + \Delta t\,v - \eta\,g,
\qquad v=(\hat{x}_1-x_t)/(1-t).
\]
The guidance strength $\eta$ is tuned per cell from
$\{0.05,0.3,0.5,1.0,2.0\}$, with larger values used at smaller $M$
where the rendering signal is weaker.

\paragraph{Persistent latents.}
For EB-FM, the per-observation latent is updated each E-step by
$\tilde{x}_i \leftarrow \alpha\,\tilde{x}_i+(1-\alpha)\,\hat{x}_i$
with $\alpha\in\{0.5,0.9\}$ tuned per cell with larger $\alpha$ at
well-determined cells and  smaller $\alpha$
at low $M/D$. The
EB-FM w/o PL ablation replaces the latent fully each iteration
($\alpha=0$).

\paragraph{DiEM-iso baseline.}
We use a tractable variant of DiEM~\cite{rozet2024learning} with the
isotropic-covariance approximation, which avoids back-propagation
through the network inside the CG solve. We use ancestral DDPM with
$N=50$ steps; the $M\times M$ system is solved by direct linear solve
when $M<16$ and by 5-iteration conjugate gradient otherwise.

\paragraph{Outer EM schedule.}
All three methods share the same training schedule: a 50-step
rendering-only warm-up of the latents, a prior bootstrap of $50{,}000$
steps on the warm-up latents, and 100 outer EM iterations. Each outer
iteration runs one E-step (one round of guided sampling per
observation) followed by one M-step ($100$ Adam inner steps with
batch 256, $\beta_2=0.999$, learning rate $3\times 10^{-4}$).

\paragraph{Evaluation.}
For each cell we generate 1000 unconditional samples from the
trained prior projected to 2D via $P^\top$, as shown in Figure~\ref{fig:two_moons_scatter}.

\paragraph{Hardware.}
All two-moons experiments were run on a single NVIDIA RTX 4090 GPU.
Each cell takes 50--120 seconds depending on $M$ and method.

\section{MNIST Experiment}
\label{app:mnist_details}

\begin{table}
\centering
\caption{MNIST: quality of the learned EB-FM prior under linear random-projection observations ($M{=}50$ and $M{=}100$ measurements per $28{\times}28$ image). Metrics are computed between $10\mathrm{K}$ unconditional samples from the prior and $10\mathrm{K}$ held-out images: EMD (Hungarian assignment), Fidelity (each generated $\to$ nearest GT), Coverage (each GT $\to$ nearest generated), each in PSNR and SSIM. Higher is better. EB-FM with persistent latents outperforms the diffusion baseline and the EB-FM ablation without persistent latents.}
\label{tab:mnist_results}
\small{
\begin{tabular}{@{}llcccccc@{}}
\toprule
& & \multicolumn{3}{c}{\textbf{PSNR} $\uparrow$} & \multicolumn{3}{c}{\textbf{SSIM} $\uparrow$} \\
\cmidrule(lr){3-5} \cmidrule(l){6-8}
\textbf{Obs.} & \textbf{Method} & \textbf{EMD} & \textbf{Fidelity} & \textbf{Coverage} & \textbf{EMD} & \textbf{Fidelity} & \textbf{Coverage} \\
\midrule
\multirow{3}{*}{50} 
& Diffusion & 13.75 & 14.08 & 14.49 & 0.59 & 0.60 & 0.62 \\
& EB-FM w/o PL & 13.75 & 14.33 & 14.48 & 0.56 & 0.58 & 0.62 \\
& EB-FM & 15.67 & 16.20 & 16.35 & 0.70 & 0.72 & 0.74 \\
\midrule
\multirow{3}{*}{100} 
& Diffusion & 14.72 & 15.36 & 15.42 & 0.68 & 0.70 & 0.71 \\
& EB-FM w/o PL & 14.39 & 15.04 & 15.55 & 0.68 & 0.70 & 0.76 \\
& EB-FM & 15.37 & 16.06 & 16.13 & 0.71 & 0.73 & 0.74 \\
\bottomrule
\end{tabular}
}
\end{table}

In order to quantify the quality of the prior learned on MNIST, we compute PSNR and SSIM metrics using the following: (1) Earth Mover's Distance (EMD) PSNR/SSIM, matching 10K generated samples with 10K held-out ground truth samples using the Hungarian algorithm, and averaging the PSNR/SSIM over all matched pairs; (2) Fidelity PSNR/SSIM, where each generated sample is compared to the closest GT samples; and (3) Coverage PSNR/SSIM, where each GT sample is compared to the closest generated sample.  Table~\ref{tab:mnist_results} compares these metrics between EB-FM, EB-FM w/o PL, and a Diffusion baseline, all using the same network architecture, posterior guidance mechanism and outer EM loop. EB-FM outperforms the baselines. 

\paragraph{Dataset and preprocessing.}
We evaluate our method on the MNIST dataset consisting of grayscale handwritten digit images of resolution $28\times28$. 
All images are normalized to the range $[-1,1]$. 
Unless otherwise stated, experiments use $10{,}000$ training images.

\paragraph{Problem setup.}
Each image $\mathbf{x}\in\mathbb{R}^{1\times28\times28}$ is observed through a linear measurement operator
\[
\mathbf{y}=A^\top \mathbf{x}+\epsilon,
\]
where $A\in\mathbb{R}^{784\times M}$ is a random Gaussian sensing matrix and $\epsilon\sim\mathcal{N}(0,\sigma_y^2 I)$ is additive Gaussian noise. 
We use $M=100 / 50$ measurements and observation noise standard deviation $\sigma_y=10^{-3}$.

For reproducibility, each training image is assigned a deterministic sensing matrix generated using the image index as the random seed. 
Columns of the sensing matrix are $\ell_2$ normalized.

\paragraph{Flow model architecture.}
The EB-FM prior is implemented using a lightweight U-Net-style convolutional architecture operating directly in image space. 
The network receives a noisy/intermediate sample $\mathbf{x}_t$ together with time $t$, and predicts a clean endpoint estimate $\hat{\mathbf{x}}_1$.

Time conditioning is implemented using Gaussian Fourier features followed by a two-layer MLP. 
The architecture consists of residual convolutional blocks with Group Normalization and SiLU activations, average-pooling downsampling, nearest-neighbor upsampling, and attention-gated skip connections. 
The base channel dimension is 16.




\paragraph{EB-FM guided sampling.}
At inference time, reconstruction begins from Gaussian noise $\mathbf{x}_0\sim\mathcal{N}(0,I)$,
and is evolved using Euler integration with $N=20$ steps. 
At each step, the flow model predicts a clean estimate $\hat{\mathbf{x}}_1=f_\theta(\mathbf{x}_t,t)$.
Measurement consistency is enforced using the residual $\mathbf{r}=A^\top\hat{\mathbf{x}}_1-\mathbf{y}$, and guidance objective $\mathcal{L}_{\mathrm{guide}} = \|\mathbf{r}\|_2^2$.
The gradient of the guidance loss with respect to the current noisy state $\mathbf{x}_t$ is used to correct the Euler update:
\[
\mathbf{x}_{t+\Delta t} = \mathbf{x}_t + \Delta t\,\hat{\mathbf{v}}_\theta - \frac{\eta}{\|\mathbf{r}\|_2} \nabla_{\mathbf{x}_t} \mathcal{L}_{\mathrm{guide}}.
\]
We use guidance strength $\eta=1.0$.

\paragraph{Persistent latent optimization (EB-FM).}
For the full EB-FM method, we maintain one trainable latent image tensor $\mathbf{z}_i\in\mathbb{R}^{1\times28\times28}$
per training sample. 
These latent images are jointly optimized together with the flow prior.
The optimization objective combines reconstruction consistency and guided-sampling consistency:
$\mathcal{L}
=
\lambda_{\mathrm{flow}}
\|\hat{\mathbf{z}}-\mathbf{z}\|_1
+
\lambda_{\mathrm{render}}
\|A^\top\mathbf{z}-\mathbf{y}\|_2^2$.
We use $\lambda_{\mathrm{flow}}=1,
\qquad
\lambda_{\mathrm{render}}=1$.

\paragraph{EB-FM without persistent latents.}
In the non-persistent variant (EB-FM w/o PL), latent tensors are not optimized parameters. 
Instead, latents are initialized from Gaussian noise and replaced directly by the guided reconstruction samples during training.

\paragraph{Diffusion baseline.}
We compare against a diffusion-based reconstruction baseline using the same backbone architecture and measurement-guidance framework. 
The diffusion baseline replaces the flow matching objective with a standard diffusion denoising objective while keeping the same forward operator, reconstruction setup, and measurement guidance mechanism.

\paragraph{Optimization details.}
All models are trained using Adam optimization. 
The flow prior uses learning rate $10^{-4}$, while persistent latent tensors use learning rate $10^{-2}$. 
Training is performed for 2000 epochs with batch size 64 and gradient clipping with maximum norm 1.0.

For EB-FM, the flow prior is periodically refined on the current latent estimates using 5 additional local training epochs after each refinement stage. 
At the end of training, the flow prior is additionally refined for 100 epochs.

\paragraph{Hardware.}
All MNIST experiments were run on a single NVIDIA RTX 4090 GPU with 24 GB of memory. The total wall-clock runtime was approximately 37 hours for experiments using the diffusion based algorithm and 13 hours for experiments using the flow matching algorithm.

\begin{table}[h]
\centering
\caption{MNIST experimental configuration.}
\label{tab:mnist_setup}
\small
\setlength{\tabcolsep}{4pt}
\begin{tabular}{lccc}
\toprule
Setting & Diffusion & EB-FM w/o PL & EB-FM \\
\midrule
Latent LR & -- & -- & $10^{-2}$ \\
Guidance steps & 100 & 20 & 20 \\
Persistent latents & No & No & Yes \\
\bottomrule
\end{tabular}
\end{table}

\section{Cryo-EM Experimental Details}
\label{app:cryoEM}

We write \(\Call{FlowSample}{f_\theta,p_0,N}\) for Euler sampling from the flow prior for \(N\) steps, starting from \(x_0\sim p_0\).
For cryo-EM, \(\mathcal{A}\) denotes the differentiable renderer for observation \(y_i\).
The operation \(\Call{RMSDAlign}{x,x'}\) rigidly aligns \(x'\) to \(x\) by minimizing RMSD using the Kabsch algorithm; this removes global rotation and translation before computing the flow update.
The backbone-continuity loss compares the current pairwise atom distances to the expected atom distances on a fixed set of local neighbor pairs:
\[
\mathcal{L}_{\mathrm{backbone}}(x)
=
\frac{1}{|\mathcal{M}|}
\sum_{(i,j)\in\mathcal{M}}
\bigl(D_{ij}(x)-D^0_{ij}\bigr)^2,
\]
where $D(x)$ is the pairwise atom-distance matrix of structure $x$, $D^0$ is computed from the initial structure, and
\[
\mathcal{M}=\bigl\{(i,j) : 0.1\,\text{\AA} < D^0_{ij} < 6.0\,\text{\AA}\bigr\}
\]
is the set of neighbor pairs determined from the initial structure.

\begin{algorithm}[H]
\caption{Cryo-EM E-step}
\label{alg:cryoem_estep}
\begin{algorithmic}[1]
\Require Observation $y_i$, renderer $\mathcal{A}$, flow model $f_\theta$, current latent $\tilde{x}_i$
\Require Harmonic prior $p_0$, expected atom distances $D^0$
\Require Candidate count $C$, unguided steps $N_u$, guided steps $N_g$
\Require Guidance strength $\eta$, latent learning rate $\rho$
\Require Weights $\lambda_{\mathrm{flow}},\lambda_{\mathrm{render}},\lambda_{\mathrm{backbone}}$

\Statex \textbf{Candidate selection with unguided sampling}
\For{$c = 1$ to $C$}
    \State $x^{(c)} \gets \Call{FlowSample}{f_\theta,p_0,N_u}$
    \State $\mathcal{L}_{\mathrm{cand}}^{(c)} \gets \|\mathcal{A}(x^{(c)})-y_i\|_1$
\EndFor
\State $c^\star \gets \arg\min_{c\in\{1,\dots,C\}}\mathcal{L}_{\mathrm{cand}}^{(c)}$
\State $x_t \gets x^{(c^\star)}$

\Statex \textbf{Guided sampling}
\For{$n = 0$ to $N_g-1$}
    \State $t \gets n/N_g$, \quad $s \gets t+1/N_g$
    \State $\hat{x}_1 \gets f_\theta(x_t,t)$
    \State $\mathcal{L}_{\mathrm{guide}} \gets
    \lambda_{\mathrm{render}}\|\mathcal{A}(\hat{x}_1)-y_i\|_1
    +
    \lambda_{\mathrm{backbone}}\|D(\hat{x}_1)-D^0\|_2^2$
    \State $\hat{x}_1 \gets \hat{x}_1-\eta \nabla_{x_t}\mathcal{L}_{\mathrm{guide}}$
    \State $z_t \gets \Call{RMSDALIGN}{\hat{x}_1,x_t}$
    \State $v \gets (\hat{x}_1-z_t)/(1-t)$
    \State $x_t \gets z_t+(s-t)v$
\EndFor
\State $\hat{x}_i \gets \hat{x}_1$

\Statex \textbf{Latent update}
\State $\mathcal{L}_{\mathrm{latent}} \gets
\lambda_{\mathrm{flow}}\|\tilde{x}_i-\hat{x}_i\|_1
+
\lambda_{\mathrm{render}}\|\mathcal{A}(\tilde{x}_i)-y_i\|_1
+
\lambda_{\mathrm{backbone}}\|D(\tilde{x}_i)-D^0\|_2^2$
\State Update $\tilde{x}_i$ with Adam using $\nabla_{\tilde{x}_i}\mathcal{L}_{\mathrm{latent}}$

\State \textbf{return} $\tilde{x}_i$
\end{algorithmic}
\end{algorithm}

\paragraph{Flow model architecture.}
For each protein, we represent the latent 3D conformation using backbone-atom coordinates, treating each backbone atom as a token with three coordinate channels. The flow prior is parameterized by a transformer network with 3 transformer blocks, hidden dimension 256, and 8 attention heads. Each transformer block consists of multi-head self-attention, a two-layer feed-forward network with expansion factor 4 and GELU activation, and layer normalization. The model further includes a learned positional embedding over backbone atoms and a sinusoidal time embedding projected to the model dimension.

\paragraph{Flow matching training objective.}
The flow model is trained on the current latent structures using a flow matching objective. Given a latent structure \(x_1\), we sample \(x_0\) from the harmonic prior and align it to \(x_1\). We then sample \(t \sim \mathcal{U}(0,1)\), construct the interpolated structure \(x_t = t x_1 + (1-t)x_0\), and train the model to predict the corresponding velocity field. The model predicts a clean structure estimate \(\hat{x}_1\), which is converted into a velocity prediction by
\[
\hat{v}(x_t,t) = \frac{\hat{x}_1 - x_t}{1-t}.
\]
The flow loss is the mean squared error between this predicted velocity and the target velocity. For numerical stability near \(t=1\), the denominator \(1-t\) is lower-bounded by \(0.05\).

\paragraph{EB-FM guided sampling.}
Guided sampling first generates \(5\) candidate samples per observation using \(3\) unguided flow evaluations. Each candidate is rendered and compared to the observed image, and we select the candidate with the lowest rendering loss. Starting from this selected candidate, we then perform guided sampling using \(10\) additional flow evaluations with guidance strength \(0.5\).

\paragraph{Differentiable renderer.}
During training and guided sampling, we use a differentiable JAX-based renderer. Each structure is first centered by subtracting the midpoint of its atomic bounding box. The centered structure is then represented as a Gaussian mixture volume with one Gaussian centered at each backbone atom. We use a Gaussian variance parameter of \(\sigma=1\) for all experiments. The renderer produces \(64 \times 64\) images with pixel size \(1.2\,\text{\AA}\), acceleration voltage \(300\,\mathrm{kV}\), defocus \(5000\,\text{\AA}\), spherical aberration \(2.1\,\mathrm{mm}\), and amplitude contrast \(0.06\) matching the parameters in the TEM simulated dataset. The image formation model includes a contrast transfer function through an astigmatic CTF model, and each rendered image is normalized to zero mean and unit variance.

\paragraph{Persistent latent optimization (EB-FM).}
The latent structures are initialized from the first frame and optimized directly with Adam using a learning rate of \(0.1\). We use an \(L1\) rendering loss with rendering-loss weight \(0.8\) and flow-guidance weight \(0.8\). The backbone-continuity weight is set to \(0.8\) for \(1a62\_A\) and \(0.3\) for the remaining proteins. Each experiment uses \(10{,}000\) frames and one image per frame. During the warmup phase, only the latent structures are updated, using the rendering loss and backbone-continuity regularization. The warmup phase is run for 50 epochs for \(1a62\_A\) and \(3rjs\_A\), and for 200 epochs for \(1egw\_C\) and \(7r79\_C\).

\paragraph{Optimization details.}
All experiments are trained for \(3000\) epochs with a main training batch size of \(512\), corresponding to \(20\) optimizer steps per epoch and \(60{,}000\) latent-structure gradient steps in total. The flow model is trained with a learning rate of \(2 \times 10^{-4}\). After warmup, the flow model is first trained on the current latent structures for 50 epochs. It is then retrained every epoch using a flow-training batch size of \(256\), corresponding to \(40\) gradient steps per flow epoch. In addition, every 300 epochs after warmup, we perform a longer flow retraining stage of 50 epochs, and at the end of training we perform one final 50-epoch flow-training stage. This results in \(153{,}600\) total flow-model gradient steps for \(1a62\_A\) and \(3rjs\_A\), and \(147{,}600\) total flow-model gradient steps for \(1egw\_C\) and \(7r79\_C\).

\paragraph{Evaluation protocol.}
For evaluation, we use unguided sampling from the trained flow model. For each protein, we generate \(10{,}000\) structures, matching the number of conformations in the ground-truth MD trajectory. The number of unguided flow sampling steps is selected separately for each protein: \(20\) steps for \(1a62\_A\), \(7\) steps for \(7r79\_C\), \(6\) steps for \(1egw\_C\), and \(20\) steps for \(3rjs\_A\). The resulting generated ensemble is then compared against the corresponding ground-truth MD ensemble using the evaluation metrics described in the main text.

\paragraph{Proteins used.}

3rjs\_A is a small, compact dynein light-chain protein, representing a relatively constrained system. 
It has 89 amino acids in an a+b two layer structure.
1a62\_A is the bacterial transcription termination factor Rho, with an HeH motif-like domain (alpha-array) and an OB-fold domain (beta-barrel), representing a moderately flexible system. 
It has 130 amino acids.
7r79\_C is a calcium-binding protein with 78 amino acids in an alpha array structure, representing an intermediate-flexibility system with moderate conformational changes. 
1egw\_C is the human myocyte-specific enhancer factor 2A. 
It has 79 residues in an alpha+beta three layer structure, and represents a more flexible system with larger structural variability. 
This selection allows us to evaluate performance across increasing levels of conformational complexity, from near-rigid structures to more dynamic and potentially multi-modal ensembles.
    

\paragraph{Hardware.}
This research was supported by the NVIDIA Academic Grant Program using a single NVIDIA H100 GPU with 80 GB of memory for each protein. The total wall-clock runtime varied by protein: \(1\mathrm{egw\_C}\) required approximately 10 hours, \(7\mathrm{r79\_C}\) required 11  hours, \(3\mathrm{rjs\_A}\) required 14 hours, and \(1\mathrm{a62\_A}\) 23 hours.

\section{Cryo-EM Ablation}
We report an ablation  where the backbone continuity loss is omitted, which significantly degrades performance. Additionally we report sensitivity to different initializations and levels of noise found in traditional cryo-EM experimental data where there exist errors in CTF and pose parameters. We run EB-FM with different levels of noise in regards to CTF and poses as seen in  Table \ref{tab:cryo_em_ablations}. 
The results show that randomized initialization helps in 3 out of 4 proteins for most distributional metrics and for the flexibility correlation metric (RMSF r) although not for the absolute flexibility metrics (pairwise RMSD and RMSF). This suggests randomized initialization improves overall distributional accuracy, though it does not by itself increase the recovered flexibility magnitude.
The results also show that small noise in pose and CTF values can also lead to better results. We hypothesize that this is caused by reducing the bias of the intentional imperfect forward model that we use.
\begin{table}[H]
\centering
\caption{Sensitivity to initialization and hyperparameters. Randomized-init initializes each persistent latent with its own random deformation of the first frame; small-error-all applies a small pose/CTF error to every particle; $p{=}0.1$ and $p{=}0.3$ apply a pose/CTF error to 10\% and 30\% of the particles, respectively; w/o backbone-loss removes the backbone continuity loss. Best per protein in bold.}
\label{tab:cryo_em_ablations}
\renewcommand{\arraystretch}{1.3}
\setlength{\tabcolsep}{7pt}
\resizebox{\linewidth}{!}{%
\begin{tabular}{llccc|ccc}
\hline
 &  & \multicolumn{3}{c|}{\textbf{Distributional accuracy}} & \multicolumn{3}{c}{\textbf{Flexibility}} \\
\cline{3-5} \cline{6-8}
\textbf{Protein} & \textbf{Method}
& \begin{tabular}[c]{@{}c@{}}EMD\\RMSD \AA\end{tabular}
& \begin{tabular}[c]{@{}c@{}}Coverage\\RMSD \AA\end{tabular}
& \begin{tabular}[c]{@{}c@{}}Fidelity\\RMSD \AA\end{tabular}
& \begin{tabular}[c]{@{}c@{}}Pairwise\\RMSD \AA\end{tabular}
& \begin{tabular}[c]{@{}c@{}}RMSF\\\AA\end{tabular}
& \begin{tabular}[c]{@{}c@{}}RMSF\\$r$\end{tabular} \\
\hline

\multirow{7}{*}{\textbf{1a62\_A}}
& Ground truth
& -- & -- & -- & 2.40 & 1.08 & -- \\
& EB-FM
& 1.62 & 1.37 & 1.27 & 1.58 & 0.77 & 0.96 \\
& EB-FM randomized-init
& \textbf{1.59} & \textbf{1.33} & \textbf{1.23} & 1.43 & 0.71 & \textbf{0.97} \\
& EB-FM small-error-all
& 1.62 & 1.39 & 1.24 & 1.46 & 0.77 & 0.95 \\
& EB-FM p$=$0.1 error
& 1.70 & 1.45 & 1.37 & 1.66 & 0.92 & 0.93 \\
& EB-FM p$=$0.3 error
& 2.06 & 1.66 & 1.78 & \textbf{2.24} & 1.40 & 0.83 \\
& EB-FM w/o backbone-loss
& 3.00 & 2.87 & 2.45 & 1.66 & \textbf{1.07} & 0.78 \\
\hline

\multirow{7}{*}{\textbf{7r79\_C}}
& Ground truth
& -- & -- & -- & 3.21 & 1.89 & -- \\
& EB-FM
& 3.19 & 2.35 & 2.35 & 1.48 & 0.91 & 0.91 \\
& EB-FM randomized-init
& \textbf{2.86} & \textbf{2.30} & \textbf{2.27} & 1.24 & 0.73 & 0.94 \\
& EB-FM small-error-all
& 2.99 & 2.40 & 2.37 & 1.32 & 0.75 & \textbf{0.94} \\
& EB-FM p$=$0.1 error
& 3.31 & 2.48 & 2.83 & 2.39 & 1.60 & 0.75 \\
& EB-FM p$=$0.3 error
& 3.91 & 2.78 & 3.23 & \textbf{2.75} & \textbf{1.68} & 0.75 \\
& EB-FM w/o backbone-loss
& 5.43 & 5.19 & 3.64 & 1.27 & 0.80 & 0.55 \\
\hline

\multirow{7}{*}{\textbf{1egw\_C}}
& Ground truth
& -- & -- & -- & 6.45 & 3.11 & -- \\
& EB-FM
& 6.22 & 4.44 & 4.19 & \textbf{4.29} & \textbf{2.55} & 0.93 \\
& EB-FM randomized-init
& 5.56 & 4.36 & 4.10 & 3.54 & 2.07 & \textbf{0.96} \\
& EB-FM small-error-all
& \textbf{4.73} & \textbf{3.87} & \textbf{3.57} & 3.64 & 2.13 & 0.96 \\
& EB-FM p$=$0.1 error
& 5.42 & 4.27 & 4.25 & 3.53 & 2.13 & 0.95 \\
& EB-FM p$=$0.3 error
& 5.27 & 4.16 & 4.14 & 4.00 & 2.42 & 0.95 \\
& EB-FM w/o backbone-loss
& 9.12 & 8.73 & 5.05 & 2.35 & 1.46 & 0.57 \\
\hline

\multirow{7}{*}{\textbf{3rjs\_A}}
& Ground truth
& -- & -- & -- & 1.79 & 0.68 & -- \\
& EB-FM
& \textbf{1.40} & \textbf{1.22} & \textbf{0.99} & 1.10 & 0.49 & 0.95 \\
& EB-FM randomized-init
& 1.43 & 1.23 & 1.02 & 1.10 & 0.49 & \textbf{0.96} \\
& EB-FM small-error-all
& 1.44 & 1.25 & 1.00 & 0.97 & 0.45 & 0.95 \\
& EB-FM p$=$0.1 error
& 1.50 & 1.27 & 1.11 & 1.26 & \textbf{0.65} & 0.95 \\
& EB-FM p$=$0.3 error
& 1.73 & 1.36 & 1.31 & \textbf{1.58} & 0.84 & 0.87 \\
& EB-FM w/o backbone-loss
& 3.25 & 3.05 & 2.72 & 0.97 & 0.57 & 0.79 \\
\hline

\end{tabular}%
}
\end{table}

\section{Cryo-EM Simulation Model}
\begin{figure}[H]
\centering
\includegraphics[width=0.6\linewidth]{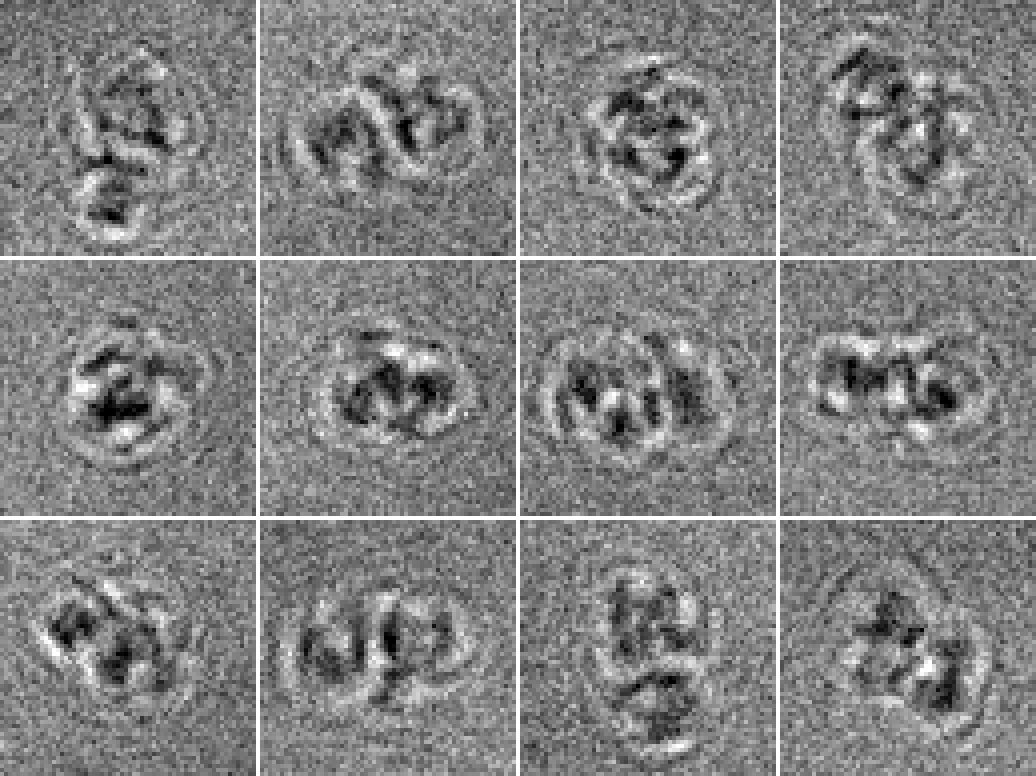}
\caption{Samples of picked particle images from our Cryo-EM TEM simulated dataset.}
\label{fig:cryo-em-samples}
\end{figure}

We follow the same simulation procedure of Rosenbaum et al \cite{rosenbaum2021inferringcontinuousdistributionatom}. Molecular dynamics ensembles of large, highly dynamic proteins are scarce, hence we used proteins of 78--130 residues. These are far smaller than the particles that can be resolved experimentally, even more so when rendered from backbone atoms only, and at standard exposures their projections carry almost no signal. We therefore simulate at an elevated dose of 100,000~e/\AA$^2$, 1,000 times higher than a reference exposure of 100~e/\AA$^2$, which raises the SNR by a factor of $\sqrt{1000}\approx 32$. Since the backbone-only rendering retains roughly half of the scattering mass, we estimate that the task is comparable in difficulty to imaging proteins about 16 times larger (roughly 1,200--2,100 residues) at 100~e/\AA$^2$.

\begin{table}[h]
\centering
\caption{Simulation parameters used for cryo-EM data generation.}
\begin{tabular}{ll}
\hline
\textbf{Parameter} & \textbf{Value} \\
\hline
Number of picked particles & 10,000 \\
Dose & $100,000\ \mathrm{e/\AA^2}$ \\
Image size & $64 \times 64$ px \\
Image pixel size & $1.2\ \mathrm{\AA}$ \\
Magnification & 25,000 \\
Defocus & $5{,}000\ \mathrm{\AA}$ \\
Accelerating voltage & $300\ \mathrm{keV}$ \\
Spherical aberration (CS) & $2.1\ \mathrm{mm}$ \\
Amplitude contrast ratio & 0.06 \\
\hline
\end{tabular}
\label{tab:cryoem_params}
\end{table}

\end{document}